\documentclass{statsoc}

\usepackage[a4paper, left = 2.5cm, right = 2.5cm]{geometry}
\usepackage[textwidth=8em,textsize=normal]{todonotes}
\usepackage{graphicx}
\usepackage{amsmath}
\usepackage{natbib, float}
\usepackage{amsfonts}
\usepackage{lscape}
\usepackage{color}
\usepackage{multirow}
\usepackage{multicol}
\usepackage{mathtools}
\usepackage{dsfont}
\usepackage{etoolbox}
\usepackage[caption = false]{subfig} 
\usepackage{enumerate}
\usepackage{comment}
\usepackage{empheq}
\usepackage{algorithm}
\usepackage{algpseudocode}

\graphicspath{{../figures/}}

\makeatletter
\patchcmd{\@makecaption}
  {\parbox}
  {\advance\@tempdima-\fontdimen2} % decrease the width!
  {}{}
\makeatother

\DeclareMathOperator{\bbu}{\boldsymbol{u}}

\DeclareMathOperator{\bX}{\hspace{-0.02cm}\boldsymbol{X}}

\DeclareMathOperator{\bbv}{\boldsymbol{v}}

\DeclareMathOperator{\bx}{\boldsymbol{x}}

\DeclareMathOperator{\bZ}{\boldsymbol{Z}}
\DeclareMathOperator{\bN}{\boldsymbol{N}}

\DeclareMathOperator{\bmu}{\boldsymbol{\mu}}

\DeclareMathOperator{\btheta}{\boldsymbol{\theta}}

\DeclareMathOperator{\0}{\boldsymbol{0}}

\DeclareMathOperator{\diag}{\mathrm{diag}}

\DeclareMathOperator*{\argmax}{arg\,max}

\DeclareMathOperator{\bV}{\boldsymbol{V}}
\DeclareMathOperator{\bU}{\boldsymbol{U}}

\DeclareMathOperator{\man}{\mathcal{M}}
\DeclareMathOperator{\nan}{\mathcal{N}}
\DeclareMathOperator{\nman}{\mathcal{N} \times \mathcal{M}_\psi}
\DeclareMathOperator{\bMd}{\boldsymbol{M}_{\partial}}
\DeclareMathOperator{\pdt}{\partial \boldsymbol{\theta}}
\DeclareMathOperator{\ret}{\mathcal{R}}
\DeclareMathOperator{\bz}{\boldsymbol{z}}
\DeclarePairedDelimiterX{\inp}[2]{\langle}{\rangle}{#1, #2}

\newcommand{\norm}[1]{\left\lVert#1\right\rVert} % define a norm structure

\title[Warped Riemannian CG]{Warped geometric information on the \\ optimisation of Euclidean functions}

\author[Hartmann et al.]{Marcelo Hartmann$^1$, Bernardo Williams$^1$, Hanlin Yu$^1$, \\ Mark Girolami$^{2}$, Alessandro Barp$^{2}$ \& Arto Klami$^1$}

\address{Department of Computer Science, University of Helsinki, Finland $^{1}$} 
\address{Department of Engineering, University of Cambridge \& The Alan Turing Institute, United Kingdom$^{2}$}

\email{marcelo.hartmann@helsinki.fi}

\begin{document}
\begin{abstract}
We consider the fundamental task of optimising a real-valued function defined in a potentially high-dimensional Euclidean space, such as the loss function in many machine-learning tasks or the logarithm of the probability distribution in statistical inference. We use Riemannian geometry notions to redefine the optimisation problem of a function on the Euclidean space to a Riemannian manifold with a warped metric, and then find the function's optimum along this manifold. The warped metric chosen for the search domain induces a computational friendly metric-tensor for which optimal search directions associated with geodesic curves on the manifold becomes easier to compute. Performing optimization along geodesics is known to be generally infeasible, yet we show that in this specific manifold we can analytically derive Taylor approximations up to $3^{\textrm{rd}}$-order. In general these approximations to the geodesic curve will not lie on the manifold, however we construct suitable retraction maps to pull them back onto the manifold. Therefore, we can efficiently optimize along the approximate geodesic curves. We cover the related theory, describe a practical optimization algorithm and empirically evaluate it on a collection of challenging optimisation benchmarks. Our proposed algorithm, using $3^{\textrm{rd}}$-order approximation of geodesics, tends to outperform standard Euclidean gradient-based counterparts in term of number of iterations until convergence. 
\end{abstract}

\section{Introduction} \label{sec:intro}

A central task in computational statistics and machine learning (ML) is defined in terms of optimization. Usually termed as {\it learning}, the goal is to find a parameter $\btheta \in \Theta \subseteq \mathbb{R}^D$ that maximises (or, equivalently, minimises) some objective function $\ell(\btheta)$. For instance, maximum a posteriori (MAP) estimation falls into this category, with $\ell(\btheta)$ corresponding to the logarithm of a posterior distribution in Bayesian Statistics. Such optimization problems are routinely solved using gradient-based methods \citep{hestenes:1952, nocedal:2006}, with stochastic versions \citep{adam} dominating the field for large-scale models such as deep neural networks and approximate second-order methods like BFGS \citep{bfgs:1989} are used for faster convergence in problems of smaller scale.

Typical optimization methods assume the objective function domain $\Theta$ to be Euclidean and they vary primarily in terms of how the search directions are specified, from direct use of gradients to various forms of conjugate gradient variants, see for example \citet{nesterov:1983}, \citet{bhaya:2004} or \citet{shanno:1978}, and how updates of those directions in combination with gradients are specified \citep{shanno:1978}. The scientific literature covers such optimization methods in great detail, with several theoretical results and practical efficiency covered in \citet{shanno:1978} and \citet{polak:1997}.

We approach the problem from the Riemannian geometry viewpoint. Rather than directly optimizing the target function $\ell$ whose domain (search space) is Euclidean, we define a new function $f$ on the target's function graph and endow the space in which the graph is immersed with a warped geometry. The domain of $f$ can now be seen as embedded Riemannian manifold with a warped metric and this is formally called warped product space, see for example \citet{oneil:1983}. For the sake of introduction, let's denote this manifold as $\man$ and its elements $\bx$ that will be made precise later on. Each point $\bx$ on the manifold encodes both $\btheta$ and the function value $\ell(\btheta)$ in a bijective manner with $\Theta$, thus the optima of $f$ on $\man$ preserves the optima of $\ell$ on $\Theta$. Because the set $\man$ is a Riemannnian manifold, we can harness the geometric information contained in the domain of $f$ and endow the optimisation routine with Riemannian tools. In the recent literature,  \citet{durui1:2022, durui:2022} and references therein have shown that optimisation on manifolds can achieve accelerated convergence rates.

For arbitrary manifolds the computational burden would increase due to the need for accounting extra Riemannian notions. For example, the notion of straight lines is replaced with geodesic paths on the $\man$ and the generalization of parallelism relies on the parallel transport operation \citep[see][]{docarmo:1992}. Those more general concepts commonly bring extra difficulties and higher computational costs as no closed-form arithmetics are usually known. However, for  particular embeddings and suitably chosen metrics it turns out that we can perform all the necessary computations for individual updates within the algorithm faster, in the sense of linear memory storage and quadratic in the number of arithmetic operations.

Our proposed algorithm follows closely the work by \cite{zhu:2020}, where the search directions in Riemannian conjugate gradient (RCG) methods \citep[see][]{sato:2021, sakai:2021, satoj:2022} and parallel transport operations are respectively replaced by a $1^{\textrm{st}}$-order geodesic approximation (retraction map) and vector transport, the latter using the idea of inverse backward retraction mapping via orthogonal projection \citep{luenberger:1972}. As also presented in \cite{zhu:2020}, these operations are of easy computation and have provided similar convergence speed performance compared to closed-form parallel transport on specific matrix manifolds \citep{absil:2008, sato:2021}. Note that these previous works account for the Riemannian manifold in simpler manner, using $1^{\textrm{st}}$-order retractions which are linear approximations of geodesic curves.

Our proposed approach builds on two key elements. First, we recast the optimisation task of a Euclidean function to the optimisation of a new function on the embedded manifold which is given by the function graph's whose embedding space is associated with a specific warped Riemannian metric. This will allow us to harness the intrinsic geometric properties of the problem to design a new optimisation algorithm. Related approaches were recently used by \citet{hartmann:2022} and \cite{Yu2023, yu:2024}, for constructing a geometric version of Markov Chain Monte Carlo samplers and Laplace's method for analytical approximations. They show that the methods induce a natural Riemannian metric-tensor that has highly desirable computational properties. For instance, we can compute its inverse metric-tensor and the Christoffel symbols in closed-form to bring down the computational costs considerably.

The second key contribution is the use of a $3^{\textrm{rd}}$-order approximations of geodesic paths as search directions. While we cannot perform efficient computation along the exact geodesics because it would require numerical solution of a system of differential equations and within it calling the metric-tensor itself several times, we show that we can construct a computationally efficient Taylor series of geodesics up to $3^{\textrm{rd}}$-order at any point on $\man$ without the need of inverting and storing full matrices at each step. \citet{Monera-2014} noted that the tangential component of the geodesic only depends on the $2^{\textrm{nd}}$-order geometry of $\man$, suggesting that both $2^{\textrm{nd}}$- and $3^{\textrm{rd}}$-order approximation are possible from theoretical viewpoint, and we are not aware of any practical algorithms that have used these approximations. As we will show, the $3^{\textrm{rd}}$-order approximation can be rewritten using only the $2^{\textrm{nd}}$-order geometry of $\man$ \citep[see also][]{song:2018} and it is not necessary to form the Hessian explicitly. Instead we directly implement its multiplication by a vector of suitable dimension. This brings down the memory cost to linear in the problem dimensionality \citep{pearlmutter:1994}. Because the approximate geodesics usually will not map back to a point in $\man$, we need to perform a retraction step to push the updated result back onto the manifold. For our case we can define a valid retraction map based on the embedding with no significant additional computation.

We evaluate the algorithm in a range of optimization tasks, covering both optimization problems with known challenging geometry and a subset of CUTE models implemented in the {\tt ADNLPmodels.jl}, Julia's package \citep{julia}. The  main goal of our experiments is to show that using the approximate geodesics as search directions reduces the number of iterations until convergence, within the scope of conjugate gradient algorithms. We show that compared to conjugate gradients with Euclidean gradient directions we observe a significant reduction, and that the proposed Riemannian conjugate gradient can be comparable to using Newton's directions that uses the inverse of the Hessian matrix in the Euclidean sense. The proposed method is also efficient in terms of overall computational speed in comparison against other methods using exact line search, but compared to methods using Hager-Zhang type of inexact line search \citep{hz:2006} it requires too many function evaluations to remain competitive in downstream tasks.

\section{Preliminaries and notation}

A set $\man$ is called {\it manifold} of dimension $D$ if together with bijective smooth mappings (at times called parametrisation) $\xi_i : \Theta_i \subseteq \mathbb{R}^D \rightarrow \man$ satisfies (a) $\cup_i \xi_i(\btheta) = \man$ and (b) for each $i$, $j$ $\xi_i(\Theta_i) \cap \xi_j(\Theta_j) \neq \emptyset$. A manifold is called a Riemmanian manifold when it is characterized by a pair $(\man, g)$ where for each $\bx \in \man$ the function $g : T_{\bx}\man \times T_{\bx}\man \rightarrow \mathbb{R}$ (called metric) associates the usual dot product of vectors in the tangent space at $\bx$ (denoted as $T_{\bx}\man$), that is $(\bV, \bU) \xrightarrow[]{g} \left\langle \bV, \bU \right\rangle_{\bx}$.  If $g$ is a non negative function we call it $\emph{Riemannian metric}$.
 
Let $(\man^m, \langle \cdot, \cdot \rangle_{\man})$ and $(\nan^n, \langle \cdot, \cdot \rangle_{\nan})$ be Riemannian manifolds of dimensions $m$ and $n$ respectively. Also let $\psi : \nan \rightarrow (0, \infty)$ be a positive and smooth function namely {\it warp function}. The product $\man \times \nan$ endowed with the Riemannian metric 
\begin{equation} \label{eq:warpm}
g = \langle \cdot, \cdot \rangle_\psi = \psi^2 \langle \cdot, \cdot \rangle_{\man} + \langle \cdot, \cdot \rangle_{\nan},
\end{equation}
is called \emph{warped product space} and denoted as $\nan \times \man_{\psi} $. Let $\man$ $=$  $\mathbb{I} \subset \mathbb{R}$ and $\nan$ $=$ $\Theta$ where $\Theta$ is the $D$-dimensional open set of $\mathbb{R}^D$ with the usual Euclidean metric. Denote as $\ell : \Theta \rightarrow \mathbb{I} \subseteq \mathbb{R}$ an arbitrary function whose graph is defined as $\Gamma_\ell = \{ (\btheta, \ell(\btheta)) : \btheta \in \Theta \}$. The \emph{canonical parametrisation} of $\Gamma_\ell$ in $\nan \times \man_\psi$ is set as $\xi : \Theta \rightarrow \Gamma_\ell \subset \nan \times \man_\psi$ where $\xi(\btheta) = (\btheta, \ell(\btheta))$. Let's denote tangent vectors at $\bx \in \Gamma_\ell$ as $\mathrm{d}\xi_{\bx}(\bbv) = \boldsymbol{M}_\partial\boldsymbol{v}$ and $\mathrm{d}\xi_{\bx}(\bbu) = \boldsymbol{M}_\partial\boldsymbol{u}$ where $\boldsymbol{M}_\partial = [\partial_1 \xi \ \ldots \ \partial_D \xi]$ stacks the tangent basis vectors associated with the canonical parametrisation and $\bbu, \bbv \in \Theta$. Then, the induced metric on $T_{\bx}\Gamma_\ell$, using \eqref{eq:warpm} is given by
\begin{align} \label{eq:indu}
    \langle \mathrm{d}\xi_{\bx}(\bbv), \mathrm{d}\xi_{\bx}(\bbu) \rangle_\psi &= \langle \boldsymbol{M}_\partial\boldsymbol{v} , \boldsymbol{M}_\partial\boldsymbol{u} \rangle_{\psi} \nonumber \\
    % &= \psi^2 \langle \bbu^\top \nabla \ell, \bbv^\top \nabla \ell \rangle + \langle \bbu, \bbv \rangle \nonumber \\
    &= \boldsymbol{v}^\top \big(I_D + \psi^2 \nabla\ell \nabla\ell^\top \big) \boldsymbol{u}^\top
    =: \langle \bbv, \bbu \rangle_{G(\bx)}
\end{align}
where $G(\bx) = I_D + \psi^2 \nabla\ell \nabla\ell^\top$ is the {\it warped metric-tensor}. From now on we will omit the argument of functions that will depend either on $\btheta \in \Theta$ or $\bx \in \Gamma_\ell$ and recall that since $\xi$ is a bijection we will only make the use of the notation $\btheta$ or $\bx$ as a variable of a function whenever the current text passage calls it necessary. Observe that the metric-tensor $G$ above has the same structural properties as the metric proposed by \cite{hartmann:2022} where the function $\psi$ plays a more general role rather than a fixed scalar value.

\section{Problem formulation and method overview}
% manifolds allow to properly define differential calculus in a coordinate-invariant way. But also I think we should emphasize that concepts in optimisations, such as “the direction of greatest rate of change”, “gradients” or the “rate of convergence”, are all coordinate-invariant concepts that depend on geometric structures, thus making Riemannian the natural mathematical setting to discuss these. In particular then I would cite these articles by Amari and so on. Finally we should add that coordinate systems do matter when we implement the algorithm for computational reasons (and we can cite your paper), but they only matter then (i.e., for computational implementation).

The notion of manifold and differentiable manifolds allow us to extend differential calculus to spaces more general than Euclidean in a coordinate-invariant (or parametrisation-invariant) manner. For example, the idea of gradient of a function as the highest rate of change at a given point becomes invariant under a differentiable manifold viewpoint. The role of the tangent space above is to do exactly this; if we were to choose a different global atlas $\tilde{\xi}(\tilde{\Theta}) = \Gamma_\ell$ representing the manifold, the tangent vector would only have a different basis but is still the same. Hence, under this property the above notion of gradient also becomes parametrisation invariant. This seems, at least, a compelling reason to perform practical optimisation procedures using notions of Riemannian geometry so that it would free us of the task of choosing a coordinate-system (parametrisation) on which the optimisation procedure theoretically behaves the best. From a computational perspective this is also important as it allows us, whenever arithmetically solvable, to choose a parametrisation so that a metric $G$ would be fully diagonal. This way, computational procedures would clearly incur faster algorithms see details in \cite{hartmann:2018} and references therein. Moreover, \citet{amari:1998}, \cite{hartmann:2018} and \cite{durui1:2022, durui:2022} observed that geometric notions can make algorithms less prone to stability issues. They can also improve the conditional number \citep{hird:2023} which in turn make them more reliable and overall lead to faster convergence rates. From now on, we will introduce the problem and formulate it from the Riemannian viewpoint. 

Consider that $\ell$ is now an objective function for which we aim to solve the maximization task 
\begin{equation}
\btheta_* = \underset{\btheta \in \Theta}{\argmax \ell(\btheta)}.
\end{equation}
We rephrase the optimisation of the function $\ell$ to a problem of maximizing a function $f : \Gamma_\ell \rightarrow \mathbb{R}$ where $\Gamma_\ell$ is an embedded manifold with the metric given in $\eqref{eq:indu}$. It will be useful later on to observe now that the embedding space $\nan \times \man_{\psi}$ endowed with the metric $g$ is the same as $\mathbb{R}^{D+1}$ with a unitary diagonal metric except for the last component of its main diagonal whose entry is the value of the warp function. The tangent basis vectors at a point in $\nan \times \man_{\psi}$ will be denote as $\bar{\bMd} = [e_1 \ \cdots \ e_{D+1}]$ where $e_i$ is the canonical basis. This way any tangent vector $\bV \in T_{\bx}(\nan \times \man_{\psi})$ can also be represented as $\bar{\bV} = \bar{\bMd} \bV$. 

Let's now specify the mapping $\Gamma_\ell \ni \bx \xrightarrow[]{f} x_{D+1}$ and since $\xi$ is a bijection between $\Gamma_\ell$ and $\Theta$ we have,
\begin{equation} 
\bx_{*} = \underset{\bx \in \mathcal{M}}{\argmax f(\bx)} \ \textrm{where} \ \bx_{*} = (\btheta_*, \ell(\btheta_*)) \ \textrm{and} \ \btheta_* = \underset{\btheta \in \Theta}{\argmax \ell(\btheta)}.
\label{eq1}
\end{equation}
This means that the first $D$ components of $\bx_* \in \Gamma_\ell$ are the same as $\btheta_* \in \Theta$. As $\Gamma_\ell$ is now the search space endowed with a geometry that is Riemannian \cite[see][for example]{docarmo:1992, docarmo:2016} we can harness its intrinsic geometric information and design an optimisation algorithm based on Riemannian concepts.
\begin{figure}[!t]
    \centering
    \includegraphics[scale = 0.3]{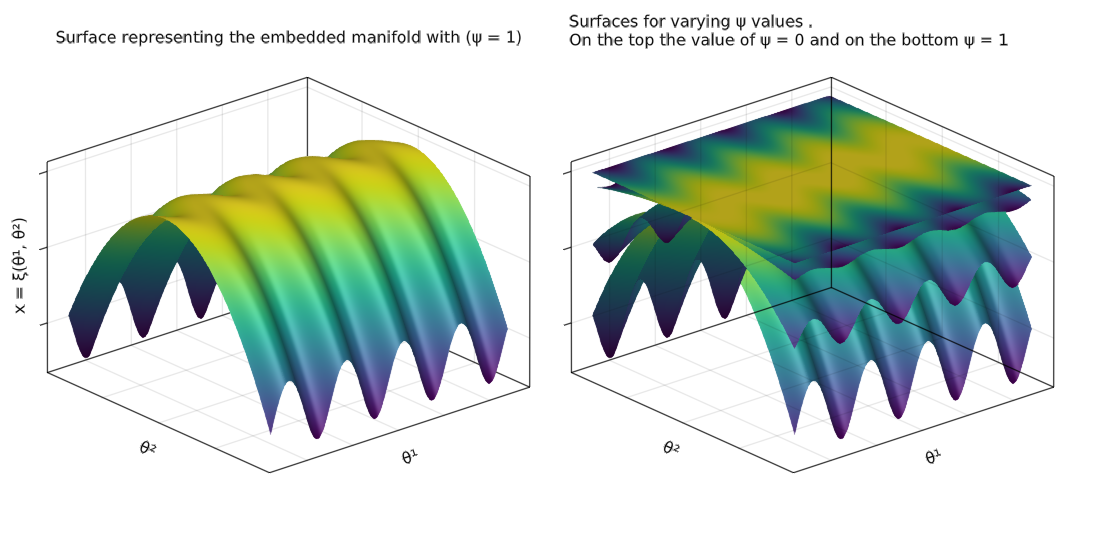}
    \caption{Visual interpretation of the domain of the functions $\ell$ and $f$. On the left panel, the plane region $(\theta_1, \theta_2) \in \Theta$ is to be understood as Euclidean. The coloured surface depicts where the function $f$ is defined on the graph of $\ell$, that is on $\Gamma_\ell$ and the ambient space as $\nman$. In this example the function $\ell(\btheta) = \log \mathcal{G}\big([\theta_1, \theta_2 + \sin(\theta_1)]|\bmu, \Sigma \big)$ where $\mathcal{G}$ denotes the Gaussian density  $\bmu = \0$ and $\Sigma = \diag(1, 0.01)$. The set $\Gamma_\ell$ has element $\bx = (\btheta, \ell(\btheta))$ and is showed on the "height" axis. This set can be understood as a embedded Riemannian manifold in the higher-dimensional space $\nman = \mathbb{R}^3$ (associated with the warped metric). On the right panel we show the behaviour of the domain of $f$ as a function of a given warp function $\psi$. As $\psi$ is closer to zero, the closer to Euclidean the set (or geometry) $\Gamma_\ell$ is.}
    \label{fig:domain-fig}
\end{figure}
Besides, note that due to the Sherman-Morrison-Woodburry identity the inverse of $G$ is fast to compute since $G^{-1}(\bx) = I_D - (\psi^2/W^2) \nabla \ell \nabla \ell^\top$
% %
% \begin{equation}
% G^{-1}(\bx) = I_D - \tfrac{\psi^2}{W^2}\nabla \ell \nabla \ell^\top.
% \end{equation}
% %
where $W = \sqrt{\psi^2 \|\nabla \ell\|^2 + 1}$. We also see that $W = \det G$, which is fast to compute.

\section{Riemannian conjugate gradient (RCG) with backward retraction} \label{sec:RCG}

The manifold $(\Gamma_\ell, g)$ characterizes all the geometric information of the domain of $f$ which can now be used to perform optimization of $\ell$ through $f$ using an iterative procedure following the general template of the RGC method presented by \cite{zhu:2020}. At each step (a) we identify a point $\bx$, a search direction and the current Riemannian gradient, (b) obtain a new $\bx'$ by optimizing the objective along a given curve passing through $\bx$ with the search direction given in (b), (c) transport the current search direction on $\bx$ to $\bx'$ and repeat the above steps. We presented the mains steps of the algorithm and our main contributions in what follows, leaving the extra technical details of the algorithm itself in the original paper of \citet{zhu:2020} and the mathematical development of this paper in the appendices.

Let $f : \Gamma_\ell \rightarrow \mathbb{R}$ be a smooth function. RCG methods ideally rely on the exponential map and parallel transport \citep[see][Section 2 and 3]{docarmo:1992}. That is, for a point $\bx_k\in \Gamma_\ell$ and a tangent vector at $\bx_k$, $\bV_{\hspace{-0.08cm} k} = \bMd \bbv \in T_{\bx_k}\Gamma_\ell$, the general form of the iterative updates is given by, 
\begin{align*}
    \bx_{k+1} &= \exp_{\bx_k} \hspace{-0.08cm}(t_k \hspace{-0.08cm}\bV_{\hspace{-0.08cm} k})\\
    \bV_{\hspace{-0.08cm} k + 1} &= \textrm{grad} f(\bx_{k+1})-\beta \mathcal{P}_{\bx_k, \bx_{k + 1}}(t_k \boldsymbol{V}_k),
\end{align*}
where $\mathcal{P}_{\bx_k, \bx_{k + 1}} : T_{\bx_k}\Gamma_\ell \rightarrow T_{\bx_{k + 1}}\Gamma_\ell$ is the parallel transport of $\bV_{\hspace{-0.08cm} k}$ along the geodesic from $\bx_k$ in direction of $\bV_k$ to $\exp_{\bx_k} \hspace{-0.08cm}(t_k\hspace{-0.08cm}\bV_{\hspace{-0.08cm} k})$, and $\textrm{grad}f$ denotes the Riemannian gradient (the natural gradient, see Appendix \ref{app:ng}). Note that the choice of the scalar $t_k$ must satisfy the Wolfe conditions in Riemannian settings \citep[see][for example]{absil:2008}. For the scalar parameter $\beta$ many choices are also possible, each of which will impact the speed of convergence of RCG \citep[see][for empirical evaluation]{sato:2021}. In practice RCG methods are difficult, both geodesics and parallel transport require solving a system of differential equations whose solution is usually computed using numerical solvers. That is why, only in the last decades, these methodologies have been used only for a few matrix manifolds \citep{absil:2008, byrne:2013} where the exponential map and parallel transport have closed-form arithmetics.

Usually, in practice, the exponential map is replaced by the \emph{retraction map} $\mathcal{R}_{\bx_k} \hspace{-0.08cm}(t_k \hspace{-0.08cm}\bV_{\hspace{-0.08cm} k})$ and the \emph{parallel transport} by the vector transport $\mathcal{T}_{\bx_k, \bx_{k + 1}}(t_k \boldsymbol{V}_k)$. In this way the iterative updates take the form,
\begin{align*}
    \bx_{k+1} &= \mathcal{R}_{\bx_k} \hspace{-0.08cm}(t_k \hspace{-0.08cm}\bV_{\hspace{-0.08cm} k})\\
    \bV_{\hspace{-0.08cm} k + 1} &= \textrm{grad} f(\bx_{k+1})-\beta \mathcal{T}_{\bx_k, \bx_{k + 1}}(t_k \boldsymbol{V}_k).
\end{align*}
From the numerical viewpoint both of these operations, when suitable defined, do not require solving a system of differentials equations. Moreover, they can alleviate the computational overhead considerably while preserving the convergence guarantees of RCG methods \citep{absil:2008, boumal:2023}. We refer to these operations as in \cite{absil:2008}, Definition 4.1.1 (page 55) and Definition 8.1.1 (page 169).
%
% \begin{defn}{Retraction} \label{def:ret}. A retraction at $\bx \in \Gamma_\ell$ is a smooth map with the following properties. Let $\ret_{\bx} : T_{\bx}\Gamma_\ell \rightarrow \Gamma_\ell$ be a retraction of $\ret$ at $\bx$ then
% %
% \begin{enumerate}
% \item $\ret_{\bx}(\0) = \bx$
% \item $D\hspace{-0.05cm}\ret_{\bx}(\0) : T_{\bx}\Gamma_\ell \rightarrow T_{\bx}\Gamma_\ell = \mathrm{id}$ is the identical map. 
% \end{enumerate}
% %
% Equivalently for any given curve defined as $c(t) = \ret_{\bx}(t\hspace{-0.07cm}\bV)$, the retraction map satisfies $c(0) = \bx$ and $\dot c(0) = \bV$. 
% \end{defn}

% \begin{defn}{Vector transport} \label{def:vect}. A vector transport between two tangent spaces $T_{\bx}\Gamma_\ell$ and $T_{\by}\Gamma_\ell$ is a map $\mathcal{T}_{\bx, \by} : T_{\bx}\Gamma_\ell \rightarrow T_{\by}\Gamma_\ell$ satisfying the following properties,  
% %
% \begin{enumerate}
% \item There exists an associate retraction $\ret$ such that $\mathcal{T}_{\bx, \ret(\bU)}(\bV)\in T_{\ret(\bU)}\Gamma_\ell$ for all $\bV, \bU \in T_{\bx}\Gamma_\ell$
% \item $\mathcal{T}_{\bx, \bx}(\bV)  = \bV$
% \item for any $a, b \in \mathbb{R}$, $\mathcal{T}_{\bx, \by}(a\bV + b\bU) = a \mathcal{T}_{\bx, \by}(\bV) + b \mathcal{T}_{\bx, \by}(\bU)$
% \end{enumerate}
% %
% \end{defn}
%
Recently, \citet{zhu:2020} have proposed a RCG where the vector transport is defined via a \emph{backward retraction map}, which is a way of measuring the displacement of two points on a manifold using tangent vectors. For general submanifolds of the Euclidean space, such as the submanifold $\Gamma_\ell$ we are working with, this is computationally feasible and fast to evaluate. They also show that by doing so, their algorithms are able to reduce wall-clock time to reach convergence (see Table 2, Section 6 in their paper).

% The retraction map is an approximation for the geodesic path to a certain degree and the vector transport is an approximation of the parallel transport operation. 

This work presents the general RCG with the inverse backward retraction that subsumes the method proposed by \citet{zhu:2020} (see Section 3, 5 and Equation (46) in their paper) and generalized by \cite{satoj:2022} (Section 4). On the top of those formulations, we propose a retraction map based on the $3^{rd}$-order Taylor approximation of the geodesic path. For the vector transport, we proposed the inverse backward retraction map. As we will show, for the embedded manifold $\Gamma_\ell$, both retraction map and vector transport will incur linear cost in memory requirements $(\mathcal{O}(D))$ and quadratic cost in the number of arithmetic operations $(\mathcal{O}(D^2))$. Following the next sections, we present the Taylor approximation, the choice of retraction based on it and the particular form of the backward retraction map. After that, we finally present the RCG optimisation algorithm using these particular tools.

\section{Third-order geodesic approximation} \label{sec:geod}

A geodesic on $\Gamma_\ell$ is a curve $\gamma : I \subseteq \mathbb{R} \rightarrow \Gamma_\ell$ that minimizes distance between two points on $\Gamma_\ell$. It generalizes the notion of straight path on flat spaces (straight lines). Equivalently, the classical $2^{\textrm{nd}}$-order derivative of these curves have only the normal vector component at each point $\gamma(t)$. That is, $\ddot \gamma (t) \in T_{\gamma(t)} \Gamma_\ell^{\perp}$ where superscript $^{\perp}$ denotes the orthogonal complement of a set.

Following \citet{Monera-2014} we compute a $3^{\textrm{rd}}$-order approximation of a geodesic by explicitly considering the parametrisation $\xi$ of $\Gamma_\ell$. Let $\gamma(t) = \xi(\btheta(t))$ where $\btheta : (0, \infty) \rightarrow \Theta$ is a curve on the chart which yields a geodesic $\gamma(t) \in \Gamma_\ell$. Recall that the exponential map is also a retraction map, $\exp_{\bx} : T_{\bx}\Gamma_\ell \rightarrow \Gamma_\ell$, which we express as $\exp_{\bx}(t\boldsymbol{V})$ $=$ $\gamma_{\bx, \bV}(t)$ where $\bV  \in T_{\bx}\Gamma_\ell$. Take a vector $\bV \in T_{\bx}\Gamma_\ell$ so that $\bV \in \mathcal{S}^D(T_{\bx}\Gamma_\ell)$ where $\mathcal{S}^D$ is the $D$-dimensional unit sphere. Then the $3^{\textrm{rd}}$-order approximation of the geodesic at $\bx = \gamma(t)$ with $t = 0$, in the direction of the tangent vector $\bV$, is given by,
\begin{align}\label{eq:geodappro}
	\tilde{\gamma}_{\bx, \bV}(t_*) = \bx + t_* \boldsymbol{V} + \frac{t_*^2}{2} \ddot \gamma(0) + \frac{t_*^3}{6} \dddot \gamma(0).
\end{align}
where $t_* \in \mathbb{R}$. From the fact that $\gamma$ is a geodesic the quadratic component of the Taylor series $Q_{\bx}(\bV) := \ddot \gamma(0) \in T_{\gamma(0)}\Gamma_\ell^{\perp}$ and so the $2^{\textrm{nd}}$-order geometry of $\Gamma_\ell$ around $\bx$ only depends on the second-fundamental form \citep[see][for example]{docarmo:2016}. \citet{Monera-2014} also observed that the tangential component of $\dddot \gamma(0) = K_{\bx}(\bV)$ only depends on the $2^{\textrm{nd}}$-order geometry of $\Gamma_\ell$. We will exploit these properties to compute a $3^{\textrm{rd}}$-order Taylor approximation of the geodesic.

We start by noting that, in general, the second derivative of a curve on the embedded manifold $\Gamma_\ell$ can be written as, 
\begin{align}
\ddot \gamma(t) = \nabla_{\dot \gamma(t)} \dot \gamma(t) + \mathbb{II}_{\gamma(t)}(\dot \gamma(t)) \bN_{\gamma(t)},
\end{align}
where $\nabla_{\bV}\hspace{-0.05cm}\bX$ denotes the covariant derivative of a tangent $\bX$ in the direction of $\bV$ \citep[see][Chapter 2]{docarmo:1992}, $\bN_{\gamma(t)}$ denotes the normal component at $\gamma(t)$ and $\mathbb{II}$ the second-fundamental form on $T_{\gamma(t)}\Gamma_\ell$ in the direction of $\dot \gamma(t)$ \citep[][Chapter 6]{docarmo:1992}. Then, if we express  $\dot \gamma(t) = \bMd \bbv$, where $\frac{\mathrm{d}}{\mathrm{d}t}\xi^{-1}(\gamma(t)) = \bbv$, the covariant derivative above can be expressed in matrices forms as 
\begin{align} \label{eq:geodeq}
\nabla_{\dot \gamma(t)} \dot \gamma(t) =  \bMd \left(
  \begin{bmatrix}
 \norm{\bbv}^2_{\Gamma^1(\gamma(t))} \\ 
 \vdots \\ 
 \norm{\bbv}^2_{\Gamma^D(\gamma(t))}
 \end{bmatrix}
 + \dot \bbv \right),
 \end{align}
where $\frac{\mathrm{d}^2}{\mathrm{d}t^2} \xi^{-1}(\gamma(t)) = \dot \bbv$. The Christoffel symbols $\Gamma^{m}(\gamma(t))$ above, assuming the natural Levi-Civitta connection,
were arranged in matrices ($D \times D)$ for easiness and mnemonic notation. Their general formulation can be found in \cite{docarmo:1992}, Chapter 2, page 56.
% %
% \begin{align}
% \Gamma^m_{i, j}(\gamma(t)) & = \tfrac{1}{2} \sum_{k = 1}^D G^{-1}_{k, m}(\gamma(t)) \big(\partial_i G_{j, k}(\gamma(t)) + \partial_j G_{k, i}(\gamma(t)) - \partial_k G_{i, j}(\gamma(t)) \big).
% \end{align}
% %
% Here $G_{i, j}^{-1}$ is the $(i, j)$ entry of the inverse of the metric-tensor $G$ and $\partial_k G_{i, j}$ is the derivative of $G_{i, j}$ with respect to the $k^{th}$ component of $\xi^{-1}(\gamma(t))$ \citep[see][for more details]{hartmann:2022, docarmo:1992}. All the indexes have the same range $i, j, k = 1, \ldots, D$. 

Since geodesics have only normal component, the coordinates of the tangent component must be the zero vector, thus for the quadratic component it holds $Q_{\bx}(\bV) = \mathbb{II}_{\bx}(\bV)\bN_{\bx}$, where $\dot \gamma(0) = \bV$ at $\bx = \gamma(0)$. Because $\Gamma_\ell$ is an embedding there is a unique normal vector at $\bx$ which we have denoted as $\bN_{\bx}$, such that it is of length one (under the warp metric) and it is orthogonal to any vector in $T_{\bx}\Gamma_\ell$. In our case this reads $\bN_{\bx} = \left( -\psi\nabla \ell/W, 1/(\psi W) \right)$ (see Appendix \ref{app:normalv} for proof).
% 
% \begin{align}
% \bN_{\bx} = \left( -\frac{\psi\nabla \ell}{W}, \frac{1}{\psi W} \right).
% \end{align}
% 

The second-fundamental form is a bilinear form defined as $\mathbb{II}_{\bx}(\bV) = - \inp{\bar{\nabla}_{\bar{\bV}} \bar{\bN}}{\bar{\bV}}_{\psi}$ where $\bar{\nabla}$ is the natural connection associated with the metric of the ambient space $\nman$(see Equation \eqref{eq:warpm}) and the vectors $\bar{\bV}, \bar{\bN}$ are natural extensions of $\bV$ and $\bN_{\bx}$. Specifically, after a long computation we obtain (see Appendix \ref{app:seconf})
\begin{align}
\mathds{II}_{\bx}(\bV) = \bbv^\top \bigg( \tfrac{2}{W} \nabla \psi \nabla \ell^\top + \tfrac{\psi}{W} \nabla^2 \ell + \tfrac{\psi}{2 W} \inp{\nabla \psi^2}{\nabla \ell} \nabla \ell \nabla \ell^\top \bigg) \bbv.
\end{align}
The computation of the cubic term of the Taylor expansion is slightly more involved as it depends on the time derivative of the second fundamental form, the normal vector and consequently depends on the geodesic equations \citep{Monera-2014}, see Appendix \ref{app:chrisgeod}. In the following, we will present the general derivative leaving the details in the Appendix \ref{app:taylor3}.
\begin{align}
 \dddot \gamma(0) 
 % &= \frac{\mathrm{d}}{\mathrm{d}t} Q_{\gamma(t)}(\dot \gamma(t)) \big|_{t = 0} \\ \nonumber
                  &= \frac{\mathrm{d}}{\mathrm{d}t} \bigg( \tfrac{2}{W} \inp{\bbv}{\nabla \psi} \inp{\bbv}{\nabla \ell} + \tfrac{\psi}{W} \|\boldsymbol{v}\|^2_{\nabla^2 \ell} + \tfrac{\psi}{2 W} \inp{\nabla \psi^2}{\nabla \ell} \inp{\nabla \ell}{\bbv}^2 \bigg) 
                  \begin{bmatrix}
                        - \tfrac{\psi\nabla \ell}{W} \\
                        \tfrac{1}{\psi W} 
                  \end{bmatrix} \Bigg|_{t = 0} \nonumber \\
                  &=: K_{\bx}(\bV). \nonumber
 \end{align}
As seen in the above equation the acceleration vector $\dot \bbv = \frac{\mathrm{d}}{\mathrm{d}t} \bbv$ appears, and once we are approximating geodesic curves the derivative $\dot \bbv$ is given by the geodesic equations. Therefore it follows that (see Appendix \ref{app:chrisgeod} for proof), 
\begin{align}
    \dot \bbv = - \mho_1 \nabla \ell + \mho_2 \nabla \psi^2.
\end{align}
where $\mho_1 = \tfrac{1}{2 W^2} \Big( 2 \langle \bbv, \nabla \psi^2 \rangle \langle \bbv, \nabla \ell \rangle + 2 \psi^2 \| \boldsymbol{v} \|^2_{\nabla^2 \ell} + \psi^2 \langle \nabla \psi^2, \nabla \ell \rangle \langle \bbv, \nabla \ell \rangle^2 \Big)$ and $\mho_2 = \tfrac{1}{2} \langle \bbv , \nabla \ell \rangle^2$.
% %
% \begin{align}
%     \mho_1 = \tfrac{1}{2 W^2} \Big( 2 \langle \bbv, \nabla \psi^2 \rangle \langle \bbv, \nabla \ell \rangle + 2 \psi^2 \| \boldsymbol{v} \|^2_{\nabla^2 \ell} + \psi^2 \langle \nabla \psi^2, \nabla \ell \rangle \langle \bbv, \nabla \ell \rangle^2 \Big)
% \end{align}
% %
% and
% %
% \begin{align}
%      \mho_2 = \tfrac{1}{2} \langle \bbv , \nabla \ell \rangle^2.
% \end{align}
% %
Thus, the $3^{\textrm{rd}}$-order Taylor approximation of a geodesic path on $\nan \times \man_\psi$ for a given $\bx$ and $\bV$ is
\begin{align} \label{eq:appgeo}
	\tilde{\gamma}_{\bx, \bV}(t_*) = \bx + t_*\bV + \frac{t_*^2}{2} Q_{\bx}(\bV) + \frac{t_*^3}{6} K_{\bx}(\bV).
\end{align}
We can now see that this final expression does not involve the inverse of Hessian or storage of it in full, but only Hessian-vector products both in $Q_{\bx}(\bV)$ and $K_{\bx}(\bV)$. Therefore the computational implementation has linear memory cost $\mathcal{O}(D)$ and it is quadratic in the number of computer operations $\mathcal{O}(D^2)$ 
% {\color{red} *is it so ? the issue seems to be only in the expression $d_t \nabla^2 \ell$ that*}. 
In the next sections we provide the choice of retraction map based on this approximation and the choice of the vector transport. 

\section{Retraction choice} \label{sec:retraction}

In Section \ref{sec:RCG}, we referred to the retraction map, that for a given point $\bx \in \Gamma_\ell$ it takes a vector $\boldsymbol{V} \in T_{\bx}\Gamma_\ell$ and maps back onto $\Gamma_\ell$. Usually the point along the approximate geodesic path \eqref{eq:appgeo} will not lie on $\Gamma_\ell$ and thus it does not satisfy the definition of the retraction map. To define a valid retraction on $\Gamma_\ell$, we proceed by applying the orthogonal projection of $\tilde{\gamma}_{\bx, \bV}$ onto $\Theta$ and using the canonical parametrisation $\xi$ to push it back onto $\Gamma_\ell$. That is, 
\begin{align} \label{eq:ret}
\mathcal{R}_{\bx}(t \boldsymbol{V}) = \xi \big( \textrm{Proj}_{T_{\xi^{-1}(\bx)} \Theta} \big( \tilde{\gamma}_{\bx, \bV}(t) \big) \big) 
\end{align}
where $\btheta = \xi^{-1}(\bx)$ and 
\begin{align} \label{eq:taylor}
\textrm{Proj}_{T_{\xi^{-1}(\bx)} \Theta} \big( \tilde{\gamma}_{\bx, \bV}(t) \big) = \btheta + t \boldsymbol{v} + \frac{t^2}{2} \boldsymbol{q} + \frac{t^3}{6} \boldsymbol{k}.    
\end{align}
The quadratic and cubic coefficients of the Taylor approximation are given by
\begin{align} \label{eq:tterms}
    \boldsymbol{q} &= - \big( \tfrac{1}{W^2} \inp{\bbv}{\nabla \psi^2} \inp{\bbv}{\nabla \ell} + \tfrac{\psi^2}{W^2} \|\boldsymbol{v}\|^2_{\nabla^2 \ell} + \tfrac{\psi^2}{2 W^2} \inp{\nabla \psi^2}{\nabla \ell} \inp{\nabla \ell}{\bbv}^2 \big) \nabla \ell \nonumber \\[0.2cm]
    \boldsymbol{k} &= - \frac{\mathrm{d}}{\mathrm{d}t} \big( \tfrac{1}{W^2} \inp{\bbv}{\nabla \psi^2} \inp{\bbv}{\nabla \ell} + \tfrac{\psi^2}{W^2} \|\boldsymbol{v}\|^2_{\nabla^2 \ell} + \tfrac{\psi^2}{2 W^2} \inp{\nabla \psi^2}{\nabla \ell} \inp{\nabla \ell}{\bbv}^2 \big) \nabla \ell \Big|_{t = 0}.
\end{align}
See Appendix \ref{app:chrisgeod} for more details. In order to show that Equation \eqref{eq:ret} is indeed a retraction map, let us show that it does satisfy the necessary properties of its definition. Define a curve $c : (0, \infty) \rightarrow \Gamma_\ell$ as, 
\begin{equation} \label{eq:geodret}
    c(t) := R_{\bx}(t\boldsymbol{V}) = \xi\left( \btheta + t \boldsymbol{v} +  \frac{t^2}{2} \boldsymbol{q} + \frac{t^3}{6} \boldsymbol{k}  \right).
\end{equation}
Evaluate this curve at $t = 0$, i.e.,  $c(0) = \xi(\btheta) = \xi(\xi^{-1}(\bx))  = \bx$ and the first property holds. For the second property we need to show that we recover $\bV = \bMd \bbv$ in the derivative $\dot c(0)$. The curve derivative is given by
\begin{align}
\dot c(t) &= \bMd \frac{\mathrm{d}}{\mathrm{d}t} \left( \btheta + t \boldsymbol{v} + \frac{t^2}{2} \boldsymbol{q} + \frac{t^3}{6} \boldsymbol{k} \right) = \bMd \left( \boldsymbol{v} + t \boldsymbol{q} + \frac{t^2}{2} \boldsymbol{k} \right)  
\end{align}
thus at $t = 0$ we have $\dot c(0) = \bMd \bbv = \bV$. Therefore we conclude that Equation \eqref{eq:ret} is a retraction map. It is also interesting to observe that the term $\mho_1$ in the Equation \eqref{eq:geodequations}, see Appendix \ref{app:chrisgeod}, equals the coefficient $\boldsymbol{q}$ and that was obtained by the projection of the normal component of the approximate geodesic curve onto $\Theta$. 
% This shows the acceleration on the chart $\Theta$ is not null and where the curved geometry information of $\nman$ plays out in $\Theta$.

\section{Vector transport as inverse backward retraction} \label{sec:vect}

The last tools necessary to complete our proposed algorithm is to define a valid vector transport. We use the inverse backward retraction map proposed by \citet{luenberger:1972} and \citet{zhu:2020} as the inverse orthographic projection. At first, this map is a projection of the difference of two points on the $\Gamma_\ell$ onto a tangent space which does not seem to characterize a vector transport. However, we still we can express it as a vector transport as follows \citep[see][for details]{satoj:2022}.

Let $\bx$, $\bz \in \Gamma_\ell$, $\Delta$ $=$ $[\bx_{1:D} - \bz_{1:D} \ \Delta \ell]$ and $\Delta \ell$ $=$ $\ell(\bx_{1:D}) - \ell(\bz_{1:D})$.  The inverse backward retraction map is defined as $\ret_{\bz}^{\textrm{bw}^{-1}}(\bx) = \textrm{Proj}_{T_{\bz}\Gamma_\ell}(\bx - \bz)$. Given $\bx$, $\bV \in T_{\bx}\Gamma_\ell$ and $\bz$ $=$ $R_{\bx}(t\boldsymbol{V})$, define vector transport operation $\mathcal{T}_{\bx, \bz} : T_{\bx}\Gamma_\ell \rightarrow T_{\bz}\Gamma_\ell$ along $R_{\bx}(t\boldsymbol{V})$ as 
\begin{align} \label{eq:vect}
    \mathcal{T}_{\bx, \bz}(\bV) &=  \mathcal{T}_{\bx, R_{\bx}(t\boldsymbol{V})}(\bV) \notag \\
    &= -\tfrac{1}{t} \ret_{R_{\bx}(t\boldsymbol{V})}^{\textrm{bw}^{-1}}(\bx) \notag \\
    % \ret_{\bx_{k + 1}}^{\textrm{bw}^{-1}}(\bx_k) &= - \frac{1}{t_k} \textrm{Proj}_{T_{\bx_k}\man}(\bx_{k + 1} - \bx_k) \\
    % &= -\tfrac{1}{t} \textrm{Proj}_{T_{R_{\bx}(t\boldsymbol{V})}\man}(\bx - R_{\bx}(t\boldsymbol{V})) \notag  \\
    &= -\tfrac{1}{t} \boldsymbol{M}_\partial \Big( \boldsymbol{M}_\partial^\top G_\psi \boldsymbol{M}_\partial \Big)^{-1} \boldsymbol{M}_\partial^\top G_\psi \Delta \hspace{0.3cm} \textrm{orthogonal projection of} \ \Delta \ \textrm{onto} \ T_{\bz}\man \notag  \\
    % &= -\tfrac{1}{t} \bMd \bigg(I - \tfrac{\psi^2(\bz_{1:D})}{W^2(\bz_{1:D})} \nabla \ell(\bz_{1:D}) \nabla \ell(\bz_{1:D})^\top \bigg)  \big[I_D \ \ \psi^2(\bz_{1:D}) \nabla \ell(\bz_{1:D}) \big] (\bx - \bz) \notag  \\
    % &= -\tfrac{1}{t} \bMd \bigg[I - \tfrac{\psi^2(\bz_{1:D})}{W^2(\bz_{1:D})} \nabla \ell(\bz_{1:D}) \nabla \ell(\bz_{1:D})^\top \ \ \tfrac{\psi^2(\bz_{1:D})}{W^2(\bz_{1:D})} \nabla \ell(\bz_{1:D}) \bigg] (\bx - \bz) \notag  \\
    % &= -\tfrac{1}{t} \bMd \bigg(\Delta_{1:D} - \tfrac{\psi^2(\bz_{1:D})}{W^2(\bz_{1:D})} \left\langle \Delta_{1:D}, \nabla \ell(\bz_{1:D}) \right\rangle \nabla \ell(\bz_{1:D}) + \tfrac{\psi^2(\bz_{1:D})\Delta \ell}{W^2(\bz_{1:D})} \nabla \ell(\bz_{1:D}) \bigg) \notag  \\
    &= -\tfrac{1}{t} \bMd \bigg(\Delta_{1:D} - \big( \left\langle \Delta_{1:D}, \nabla \ell(\bz_{1:D}) \right\rangle - \Delta \ell \big) \tfrac{\psi^2(\bz_{1:D})}{W^2(\bz_{1:D})} \nabla \ell(\bz_{1:D}) \bigg).
\end{align}
Observe that when $\psi \rightarrow 0^+$ we have the retraction $R_{\bx}(t\boldsymbol{V})$ $=$ $[\bx_{1:D} + t\boldsymbol{v} \ 0]$ thus the vector transport becomes $\mathcal{T}_{\bx, R_{\bx}(t\boldsymbol{V})}(\bV)$ $=$ $\tfrac{1}{t} [\bx_{1:D} + t\boldsymbol{v} - \bx_{1:D} \ 0] = [\bbv \ 0]$, that is, we recover the Euclidean parallelism on $\Gamma_\ell$. See also Appendix \ref{app:orthop} for orthogonal projection, and how to recover a vector $\bbv$ when given $\bV$.

\section{The novel RCG algorithm} \label{sec:nrcg}

After having obtained Equations \eqref{eq:ret} and \eqref{eq:vect} as the retraction map and the vector transport, we now propose a new RCG algorithm that optimises the function $f$ and therefore the function $\ell$. This is presented in Algorithm \ref{conj_grad_vect}. In the Step 1, we set the initial tangent vector $\boldsymbol{V}$ as the Riemannian gradient. This can be recalled as the natural gradient \citep{amari:1998}; see Appendix \ref{app:ng} for details. The Step 3, consists in the optimisation of a unidimensional composite function $c(t)$. In this function observe that the gradient $\nabla \ell$ and the Hessian-vector product $\nabla^2 \ell \bbv$ do not need to be recomputed as they can be repeatedly retrieved from the cache memory at every iteration inside this inner optimisation. Here we call the Julia package {\tt Optim.jl} to perform such univariate optimisation of the function $c(t)$. 

After the optima $t_k$ of $c(t)$ has been found, the Steps 5-8 compute $\bx_{k + 1}$, $s_k$, $\beta_k^{\textrm{DY}}$ (see Equation \eqref{eq:betady}) and $\bV_{k + 1}$. Particularly, in Steps 6-7, the computation of $s_k$ and $\beta_k^{\textrm{DY}}$ involves dot-products at the tangent spaces and these can still be further simplified, see Appendix \ref{app:comptan}. In Step 8, we do need to compute the Riemannian gradient at the new point $\bx_{k + 1}$ and the vector transport of $\bV_k$ from $\bx$ to $\bx_{k + 1}$ along $\mathcal{R}(\bV_{k}t_k)$ to set the update $\bV_{k + 1}$. However all those computations do not add more than $\mathcal{O}(D)$ memory load and $\mathcal{O}(D^2)$ arithmetic operations.
\begin{figure}[t!]
\centering
  \begin{minipage}{0.9\linewidth}
 \begin{algorithm}[H] 
    \caption{Riemannian CG method with inverse backward retraction}
    \label{conj_grad_vect}
    % Let $0<c_1<c_2<1$ parameters and $x_0\in \man$ initial point.
    Let $\bx_0\in \man$ initial point \vspace{0.0cm}
    \begin{algorithmic}[1]
        \State Obtain $\bV_{\hspace{-0.00cm} 0} = \textrm{grad} f(\bx_0) = \bMd G^{-1}(\bx_0)\nabla \ell(\xi^{-1}(\bx_0))$ \vspace{0.0cm}
        \For{$k = 0, \ldots, K$} \vspace{0.0cm}
        \State Define $c(t) = (f \circ \ret_{\bx_k})(t\hspace{-0.05cm}\bV_{\hspace{-0.05cm} k})$ \vspace{0.0cm}
        \State Compute $t_k = \underset{t \in (0, \infty)}{\argmax c(t)}$ (exact line search). 
        \State Compute $\bx_{k+1} = \ret_{\bx_k}(t_k\hspace{-0.05cm}\bV_{\hspace{-0.05cm} k})$ \vspace{0.0cm} 
        \State Compute 
        $s_k = \min \left(
        1, \frac{\|\bV_{\hspace{-0.05cm} k}\|_{\bx_k}}{\| \mathcal{T}_{\bx_k, \bx_{k + 1}} \hspace{-0.02cm} (\bV_{\hspace{-0.05cm} k}) \|_{\bx_{k+1}}}
        \right) $ \vspace{0.0cm} 
        \State Compute $\beta_k^{\textrm{DY}}$ using Dai-Yaun formula. See Equation \eqref{eq:betady} \vspace{0.0cm} 
        \State Compute $\bV_{\hspace{-0.05cm} k + 1} = 
        \textrm{grad} f(\bx_{k+1}) - \beta_k s_k \mathcal{T}_{\bx_k, \bx_{k + 1}} \hspace{-0.02cm} (\bV_{\hspace{-0.05cm} k})
        $ \vspace{0.0cm} 
        \EndFor
    \end{algorithmic}
\end{algorithm}
\end{minipage}
\end{figure}
\begin{align} \label{eq:betady}
    \beta^{\textrm{DY}}_k &= \frac{ \| \textrm{grad}f(\bx_{k + 1}) \|^2_{\bx_{k + 1}} } {s_k \langle \textrm{grad}f(\bx_{k + 1}), \mathcal{T}_{\bx_k, \bx_{k + 1}}(\bV) \rangle_{\bx_{k + 1}} - \langle \textrm{grad}f(\bx_k), \bV_{\hspace{-0.05cm}k}, \rangle_{\bx_k} } 
\end{align}

The convergence of the Algorithm \ref{conj_grad_vect} is guaranteed by the fact we have a valid retraction map and the value $t_k$ at Step 4, with exact line search also satisfies the Wolfe conditions. This implies convergence to a stationary point where the Riemannian gradient is null. See for example \citet{sakai:2021}, \citet{zhu:2020} (Section 4) and the generalization of the methodological proofs in \citet{satoj:2022}. An important question in the proposed algorithm is whether we can use variants of the scalar value $\beta^{\textrm{DY}}_k$ analogous to the Euclidean cases. For example, \citet{satoj:2022} (Equations 4.10-4.12 in that paper) requires the vector transport of the gradient to the new point on the manifold. Unfortunately, we cannot apply the vector transport defined within their proposed algorithm. There is no guarantee that by plugging $\textrm{grad} f(\bx_k)$ into Equation \eqref{eq:ret} instead of $\boldsymbol{V}_k$, it will end up at the same point on the manifold if we had used $\boldsymbol{V}_k$. Therefore it derails the use of Equation \eqref{eq:vect}. It would be possible to use an approximation of the parallel transport as noted by \cite{calin:2014} (page 253, problem 8.3) but this is left for future work. 
% Solving differential equations to perform exact parallel is possible but then the low cost computation feature of the Algorithm \ref{conj_grad_vect} would be lost making it unviable in practice. 

\section{Experiments} \label{sec:experiments}

In this section we conduct experiments using three different sets of of functions defined first entirely on Euclidean spaces with varying dimensionality $D$. For each function and dimension $D$, we perform optimisation using Algorithm \ref{conj_grad_vect} with the retraction map based on the $3^{\textrm{rd}}$-order approximation of the geodesic. As our choice of $\psi$ involves two scalar values $\alpha$ and $\sigma$ (see Appendix \ref{app:warpf} for details) we perform some preliminary runs in low dimension ($D = 2)$ to check convergence and performance. After that the selected values were $\alpha = 2$ and $\sigma = 5 \times 10^2$ for all the following experiments with three different set of model examples. 

We compare the RCG method, labelled as "RCG (ours)", with two other conjugate gradient methods, differing in the choice of the search direction and the line-search method used for determining the step size $t_k$. The first method, denoted as "CG-exact (ours)", uses standard Euclidean gradients and exact-line search when $\psi(\bx) = 0 \ \forall \bx$. This corresponds to the Euclidean version of Dai-Yuan conjugate gradient \citep{dai:1999}. We use here our own implementation, therefore the approximate geodesics are naturally replaced by straight lines in the Euclidean sense. The retraction simply becomes $\mathcal{R}_{\bx}(t\bV) = \xi(\xi^{-1}(\bx) + \bbv t)$ and the Riemannian gradient becomes the classical Euclidean gradient. The vector transport reduces to the Euclidean parallelism (see Section \ref{sec:vect}). This comparison directly quantifies the value of considering Riemannian search directions within our own implementation routine. 

The other two methods use the classical conjugate gradient method and purely Newton's directions as the search directions. Both are already implemented in the Julia package {\tt Optim.jl} and {\tt LineSearchers.jl}. Both of these methods, however, are set to use Hager-Zhang type of inexact line search as implemented in {\tt LineSearchers.jl}. We denote these as "CG-inexact" and "ND-inexact" in the upcoming plots and experiments. These methods are representative of the practical optimization algorithms used, and hence help understanding the possibilities and limitations of the proposed method.

For the first and second sets of models we know the optima $\bx_*$ beforehand, so that the stopping criteria for these models under test will be given by $|f(\bx_*) - f(\bx_k)| \leq 1 \times 10^{-16}$ (machine precision), and in all runs we set the maximum number of iterations to be $10000$. For the third set of models, we do not know the optima. The stopping criteria is then set to be the absolute function difference between consecutive points as $\Delta f = |f(\bx_{k + 1}) - f(\bx_k)| \leq 1 \times 10^{-16}$ or $\| \textrm{grad} f(\bx_{k + 1}) \| \leq 1 \times 10^{-7}$ with maximum number of iterations equal to 4000. In the next sections we specify the models and the practical experiments. \\

\noindent
\textbf{Notes on computational implementation :}
% \paragraph{Notes on computational implementation :}
The main cost in formulation of the RCG method presented lies in the number of arithmetic operations in the Hessian vector products. Except that for the time derivative of the Hessian vector product, we use automatic differentation (AD) \citep{ad:2019} via the package ${\tt ForwardDiff.jl}$. As mentioned in Section 5 of \citet{ad:2019}, when AD is carefully implemented it only increases the computational complexity by a small factor. This is also the reason we use the package {\tt ADNLPmodels.jl}, it allowed us to implement the RCG method to some models from the CUTE library using AD tools. Not all models in CUTE library are implemented in {\tt ADNLPmodels.jl} (personal communication with package developer PhD. Tangi Migot from Polytechnique Montr\'{e}al).

\subsection{The D-dimensional squiggle probability model} \label{sub:squiggle}

 The squiggle probability distribution has expression $\ell(\btheta)$ $=$ $\log \mathcal{G}\big([\theta_1, \theta_2 + \sin(a \theta_1)$, \ldots, $\theta_D + \sin(a \theta_1)]|\0, \Sigma \big)$ with parameters $a > 0 $ and $\Sigma$ positive-definite (PD) matrix, where $\mathcal{G}(\cdot|\bmu, \Sigma)$ denotes the multivariate Gaussian density function with parameters $\bmu$ and $\Sigma$. The squiggle function can have the shape of a thin sine function that concentrates its probability density around a zig-zag region, producing narrow uphill curved region towards its unique global maximiser $\btheta_* = \0$ with $\ell(\btheta_*) = - \frac{D}{2} \log(2\pi) - \frac{1}{2} \log \det(\Sigma)$. The PD matrix controls the orientation and how much thin the sine-shaped function can be. This effect is more pronounced with large values of $a$. We set these parameters to $a = 1$ and $\Sigma = \diag(30, 0.1,  \ldots, 0.1)$. As initial value for this function we set $\btheta_0 = (10, 10, \ldots, 10)$ to be far away from the maxima, this mimics real practice where we do not know the maximizer beforehand.
 
\begin{figure}[!t]
    \centering
    \hspace{-0.6cm}
    \includegraphics[scale = 0.21]{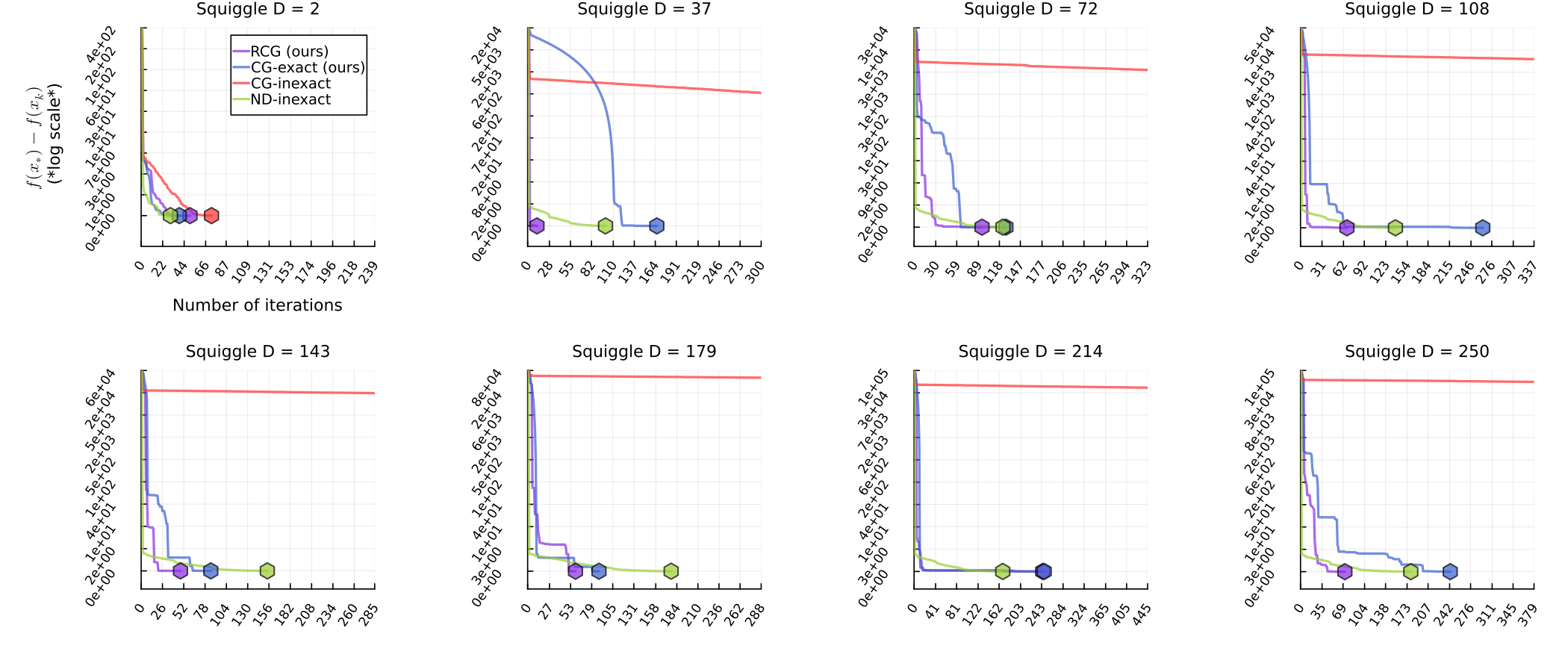}
    \caption{Number of iterations until convergence for a variety of dimensions using the squiggle model. The RCG (ours) in Algorithm 1, presents, in general, faster convergence than the, CG-exact (ours) and CG-inexact. Both RCG (ours) and CG-exact (ours), are comparable to the ND-inexact that also converges fast (in number of iterations) for all cases when compared to CG-inexact. Here the dimension is taken up to $D = 250$ 
    % since the computational wall-clock time to perform RCG starts to be larger but stabilizes 
    (See Appendix \ref{app:comp_cost} for extra information).}
    \label{fig:squiggle}
\end{figure}

 In Figure \ref{fig:squiggle}, we display the trace of three different optimisation routines as measure of computation effort (iterations) against accuracy measure (function discrepancy with its known maxima at each iteration until the maxima). This is the case again because the maximizer is known and therefore the function maxima. We also consider varying dimensionality in the experiments and plot the traces for each dimension. This experiment shows that RCG (ours) improves the number of iterations until convergence when compared to CG-inexact and it is a competitor to ND-inexact. Observe that, RCG (ours) and the CG-exact (ours) use exact line search and the other two methods, via Julia's implementation use inexact line search. For the latter cases, the Wolfe conditions may be too conservative with the step-size $t_k$ due to the "zig-zag" shape of the Squiggle function, leading to a slower convergence of the CG-inexact. The ND-inexact counterbalance this problem since the Newton's direction might be nearly orthogonal to the gradient direction. 

\subsection{The generalized Rosenbrock function} \label{sub:rosenbr}

The Rosenbrock function, $\ell(\btheta) = \sum_{i = 2}^D - b (\theta_i - \theta^2_{i - 1})^2 - (a - \theta_{i - 1})^2$, has been widely used in benchmark tests for numerical optimisation problems
\citep{rosenbr:1960, kok:2009}. For $a = 1$ and $b = 100$ its surface landscape is quite difficult. There is one global maxima in $\btheta_* = (1, \ldots, 1)$ with $\ell(\btheta_*) = 0$ and one local maxima at $\btheta_* = (-1, \ldots, 1)$ with $\ell(\btheta_*) \approx -3.99$. The global maximiser lies in a very narrow uphill region which makes the optimisation harder. The starting point for the optimisation routines is set to $\btheta_0 = (-5, 5, \ldots, (-1)^D 5)$ to make the task harder as this function has been studied in the range $-2.048 \leq \theta_i \leq 2.048$ \citep{franca:2020}. This function is not log-concave in its entire domain $\Theta$.

\begin{figure}[!t]
    \centering
    \hspace{-0.6cm}
    \includegraphics[scale = 0.21]{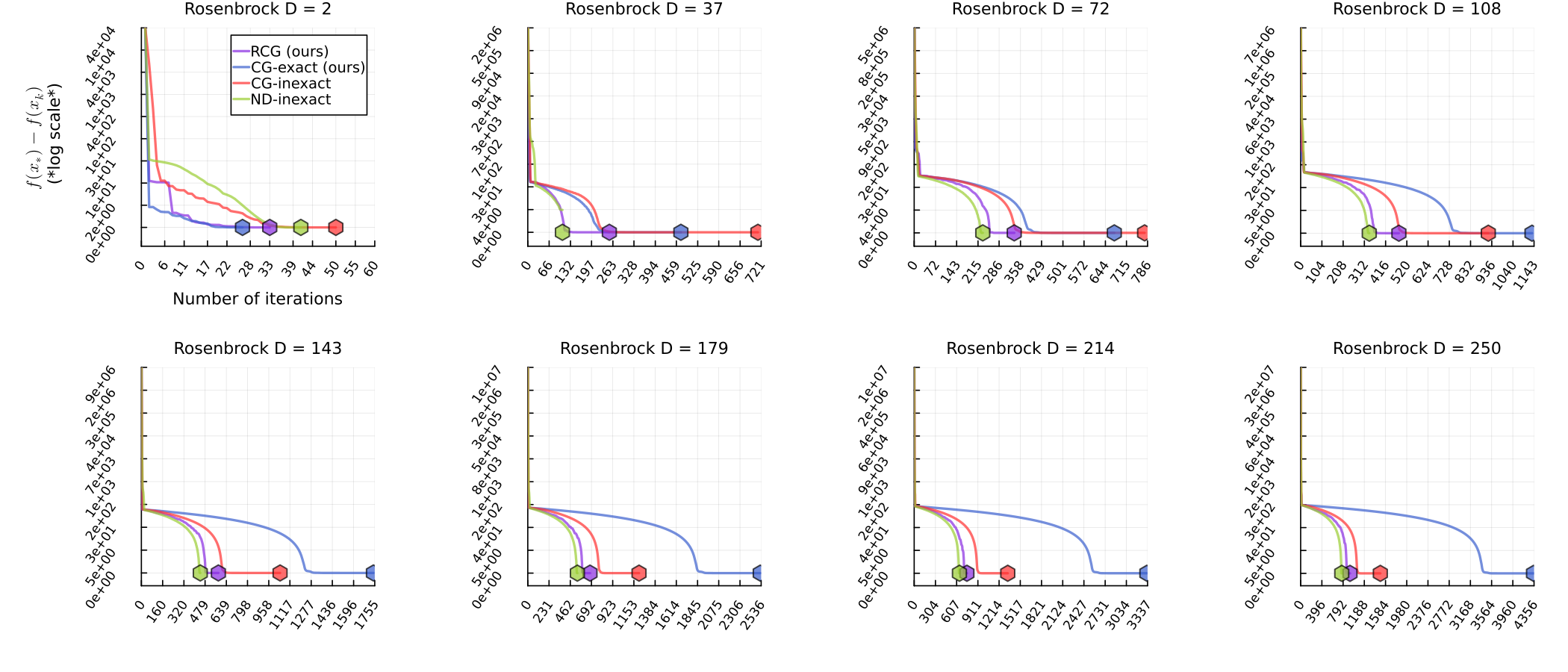}
    \caption{This figure show the number of iterations until convergence considering the Rosenbrock function with varying dimensions $D$. The parameters of the function were set to be $a = 1$ and $b = 100$ as it is usually done in benchmark settings. The RCG (ours), tends to converge faster than the CG counterparts and slower when compared to ND-inexact. See also Appendix \ref{app:comp_cost} for other types of computational cost and discussions.}
    \label{fig:rosenbr}
\end{figure}

For this set of models we plot the same measure of computation effort and accuracy as in the previous example. Figure \ref{fig:rosenbr} displays the performance of the RCG (ours), CG-exact (ours), CG-inexact and ND-inexact for variety of varying dimensions for the Rosenbrock model. In this case we also observe an improvement of RCG (ours) as the number of iterations until the discrepancy measure reaches the stopping criteria above as it decreases faster than the CG-exact (ours) and CG-inexact. The RCG (ours) is not faster than ND-inexact. Please, observe that the RCG (ours) is a first-order Riemannan optimisation scheme, not second. The analogous Riemannan second-order optimisation routine would use the Riemannian Hessian, see \citet{boumal:2023}.

\subsection{A test-set of the CUTE library}

The third and last set of models comprise some from the library CUTE (The Constrained and Unconstrained Testing Environment)\footnote{\tt www.cuter.rl.ac.uk//mastsif.html} implemented in Julia's package {\tt ADNLPmodels.jl} \footnote{\tt https://github.com/JuliaSmoothOptimizers/ADNLPModels.jl}. This subset of models can be found online\footnote{{\tt www.cuter.rl.ac.uk/Problems/classification.shtml}} and has classification SUR2-AN-V-0 (unconstrained search space). The models here chosen under this classification have IDs respectively given by EXTROSNB, CHNROSNB and GENROSE. We also vary the dimensionality of each model and use the same dimensions as in the previous tests. For all these models the initial value $\btheta_0 = \btheta_{0, M_D} + (-5, 5, \ldots, (-1)^D 5)$ where the value $\btheta_{0, M_D}$ is a quantity obtained from the model type and its respective dimension inside the library {\tt ADNLPmodels.jl}. 

% In this cases we do not know the optima and the value of the function at it, so that the stopping criteria was given by 
% the absolute function difference between consecutive points as $\Delta f = |f(\bx_{k + 1}) - f(\bx_k)| \leq 1 \times 10^{-16}$ or $\| \textrm{grad} f(\bx_k) \| \leq 1 \times 10^{-7}$ with maximum number of iterations equal to 2000.

\begin{figure}[!t]
    \centering
    \hspace{-1.0cm}
    \includegraphics[scale = 0.22]{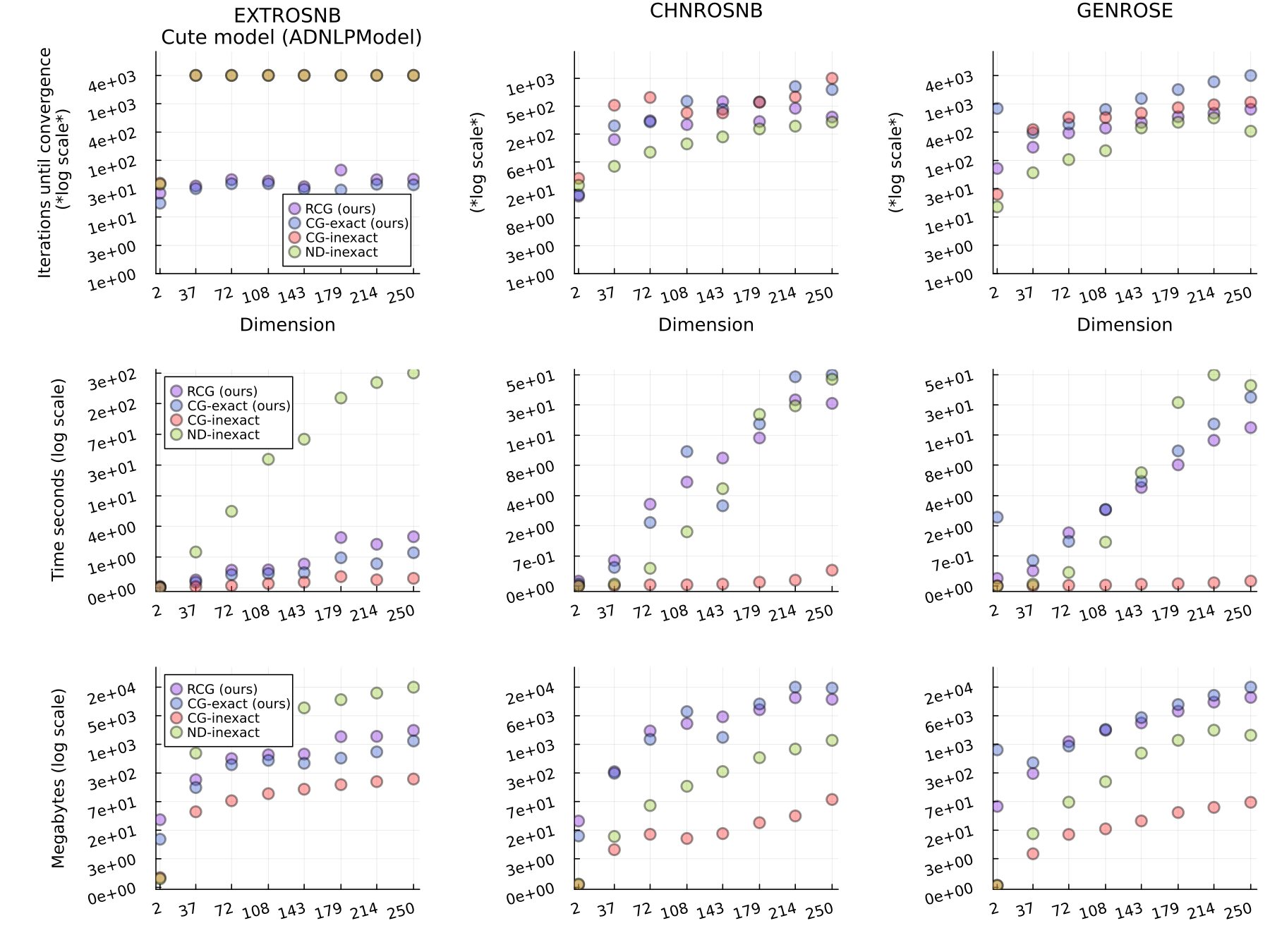}
    \caption{This figure depicts the convergence of the RCG (ours), CG-exact (ours), CG-inexact and ND-exact in the test set of problems from the CUTE library. The RCG (ours) method generally reaches smaller number of iterations to satisfy the stopping criteria when compared to CG counterparts. The ND-inexact achieves smallest number of iterations in comparison to RCG (ours) and CG counterparts for the second and third CUTE models for all dimensions. The computational cost in wall-clock time and memory requirement is generally larger for the RCG (ours) implementation.}
    \label{fig:cute}
\end{figure}

In Figure \ref{fig:cute} we display the performance of the RCG (ours), CG-exact (ours), CG-inexact and ND-inexact on this subset of models from the CUTE library. For the model EXTROSNB, we observe the performance of RCG (ours) is aligned with CG-exact (ours) and both are faster than CG-inexact and ND-inexact. In this case, the choice of parameters $(\alpha, \sigma^2)$ may be far from optimal, but the RCG (ours) was still able to be in pair with the CG-exact (ours). For the last two types of models, the RCG (ours) shows smaller number of iterations to achieve the stopping criteria when compared to the CG-exact (ours) and CG-inexact. In relation to ND-exact the RCG (ours) is not faster, achieving the stopping criteria after ND-inexact would do so. In relation to the wall-clock time and memory consumption the RCG (ours) generally dominates when compared to Julia's implementation. The wall-clock time and memory load of RCG (ours) is in line with CG-exact (ours). For these models and the chosen quality measure the CG-inexact is actually the fastest in all cases, but we note that there are likely significant differences in implementation quality when comparing our proof-of-concept implementation with established and highly optimized software packages.

% {\color{red} Here we argue that this is a question of professional computer implementation of the RCG optimisation algorithms. In Julia language, the package {\tt Optim.jl} has 103 contributors\footnote{{\tt https://github.com/JuliaNLSolvers/Optim.jl/}} and the package {\tt LineSearches.jl} has 18 contributors\footnote{{\tt https://github.com/JuliaNLSolvers/LineSearches.jl}}, this makes the comparison of our implementation and Julia's packages not all fair}.

% {\color{red} In relation to wall-clock time and time-consumption, we have additionally highlight the wall-clock time and memory consumption of the three methods, to transparently communicate the per-iteration overhead of RCG and NCG compared to the gradient CG method. In this quality measure the gradient CG is actually the fastest in all cases, but we note that there are likely significant differences in implementation quality when comparing our proof-of-concept implementation with established and highly optimized software packages.}

%

\section{Discussion and concluding remarks} \label{sec:conclusions}

% {\color{red} \cite{guy:2002} \cite{fed:2023} \cite{Yu2023} \cite{yu:2024} talk about the behaviour of the metric a bit}
We presented an alternative way to optimise a function originally defined on Euclidean space
by harnessing the geometry of the function's graph embedded in the warped product space. Despite the fact that Riemannian geometry requires extra concepts and therefore additional computational load, we were able to present a Riemannian version that has memory costs considerably smaller when compared to a complete Riemannnian procedure. Moreover, despite the fact of a large algebraic development of the RCG method presented, the exploitation of the natural geometry of the problem opens up a new venue of approaches for optimisation problems. We argue that this is important due to the following reason. In Riemannian geometry the choice of the metric tensor $g$ is free as long as it satisfies given properties. However the theory itself does not guide us on how to choose it. For example, in Statistical inference, the {\it Fisher information matrix} has been used as a metric (or pre-conditioner) in optimisation problems \citep{amari:1998} as lower-bound for the variance of unbiased estimators \citep{leh:2003, schervish:2012}. Even for already known manifolds and given metrics, the choice of optimal metrics for application problems is an open problem. See for example \citet{titsias:2023} and \citet{lan1:2016}.

Besides, experiments indicate that the proposed method improves the optimisation routine apart from its computational costs when compared to its Euclidean counterparts. This was possible due to the geodesic approximation by the Taylor expansion until 3rd-order on the function’s graph $\Gamma_\ell$ whose embedded space is associated with a warped geometry. Other variants of the RCG method such as the Riemannian version of Hager-Zhang \citep{satoj:2022} could be proposed whenever another vector transport operation would be available, though such choices are not obvious yet. 

Although the approach seems very attractive, selecting optimal values for $\alpha$, $\sigma$ is clearly not straightforward. However, compared to the previous approaches and the choice of metrics, this work is presented with larger generality, which allow us to set a warping function and thus less restrictive if for it a fixed valued would be chosen \citep{hartmann:2022, Yu2023}. Future work aims in the automatic choice of warp function $\psi$ and vector transport $\mathcal{T}$ with the goal of improve convergence speed of the Algorithm 1. Other variants of the Riemannian conjugate gradient presented here are also interesting since they may involve other update rules which potentially improves the algorithm in the lines of \citet{satoj:2022}.

Despite of the unequal match of Julia's libraries and our computer implementations, the RCG method proposed does not require the storage and inversion of any full matrix in principle. However, developing practical optimization libraries that would robustly solve general optimization problems efficiently remains as future work. One possible direction is to implement this new methodology within the Julia library {\tt manopt.jl}. This has already been pointed out by Associate Professor Ronny Bergmann in personal communication.

Observe that the induced metric-tensor on the tangent space can also be seen as a pull-back metric tensor so that it could be plugged into geometric sampling techniques, such as Riemann manifold Hamiltonian Monte Carlo (RMHMC) \citep{girolami:2011}, Lagrangian Monte Carlo (LMC) \citep{lan:2015} and Manifold Metropolis-adjusted Langenvin Dynamics \citep{txifara:2014} to cite a few. 
\\ 

% {\color{red}TODO: I think we need to here say something about the experiments and practical limitations. "In this work we evaluated the method in a range of benchmark problems and showed that it achieves convergence rates comparable to Newton's method, despite ..., and that already a proof-of-concept implementation achieves also similar total computation speed. However, developing practical optimization libraries that would robustly solve general optimization problems efficiently remains as future work.}

\noindent
\textbf{Acknowledgements}. This research is supported by the Academy of Finland grants 348952  (CORE), 345811 (ERI) and the Flagship programme Finnish Center for Artificial Intelligence (FCAI). Mark Girolami is supported by EPSRC grant numbers EP/T000414/1, EP/R018413/2, EP/P020720/2, EP/R034710/1, EP/R004889/1, and a Royal Academy of Engineering Research Chair.

% \section{Appendix} \label{app} 
\appendix \label{app}

\section{Simplification of dot-products on tangent spaces} \label{app:comptan}

Recall that for a point $\bx \in \Gamma_\ell$ and tangents expressed as $\bV = \bMd \bbv, \bU = \bMd \bbu \in T_{\bx}\Gamma_\ell$,  in the parameterization $\xi$, the inner product at $\bx$ is given by $\left\langle \bV, \bU \right\rangle_{\bx} = \left\langle \bbv, \bbu \right\rangle_{G(\bx)}$. We use that fact that $\Gamma_\ell$ is the embedded manifold to simplify the computations. The following inner products are given by

\begin{align*}
    \left\langle \bU, \bV \right\rangle_{\bx} =& \left\langle \bbu, \bbv \right\rangle_{G(\bx)} = \bbu^\top \big(I_D + \psi^2 \nabla \ell \nabla \ell^\top \big) \bbv = \left\langle \bbu, \bbv \right\rangle + \psi^2 \left\langle \nabla \ell, \bbu \right\rangle \left\langle \nabla \ell, \bbv \right\rangle
\end{align*}
and
\begin{align*}
    \left\langle \textrm{grad} f(\bx), \bV \right\rangle_{\bx} =& \left\langle \frac{\nabla \ell}{W^2}, \bbv \right\rangle_{G(\bx)} = \frac{\nabla \ell^\top}{W^2} \big(I_D + \psi^2 \nabla \ell \nabla \ell^\top \big) \bbv = \left\langle \nabla \ell, \bbv \right\rangle.
    % =& \left\langle \frac{\nabla \ell}{W^2}, \bbv \right\rangle + \frac{\psi^2}{W^2} \norm{\nabla \ell}^2 \left\langle \nabla \ell, \bbv \right\rangle \\
    % =& \left\langle \frac{\nabla \ell}{W^2}, \bbv \right\rangle + \frac{W^2 - 1}{W^2} \left\langle \nabla \ell, \bbv \right\rangle \\
    % =& \left\langle \nabla \ell, \bbv \right\rangle.
\end{align*}
If $\bV = \textrm{grad}  f(\bx)$ then $\norm{\textrm{grad} f(\bx)}^2_{\bx} = \norm{\nabla \ell}^2/W^2$.

\section{Derivation of the Riemannian gradient as the Natural gradient} \label{app:ng}

Let $\bx \in \Gamma_\ell$ such that $\xi (\btheta)= \bx$ with $\btheta \in \Theta \subseteq \mathbb{R}^D$ (the chart). Moreover, let $f\in C^\infty(\Gamma_\ell)$. By definition the Riemannian gradient is the vector in the tangent space $\textrm{grad} f \in T_{\bx} \Gamma_\ell$, $\textrm{grad} f(\bx) = (\textrm{grad} f)^ie_i $ such that for a given $\bV \in T_{\bx}\Gamma_\ell$ it holds $df(\bV) = \langle \bV, \textrm{grad} f\rangle$. Recall that the base of the tangent space is $e_i = \frac{\partial}{\pdt_i}$. Then the differential is $df(\bV) = g_{\bx}(\bV, \textrm{grad} f)$. By definition $df(\bV) = \bV(f)$ and we get $\bbv^\top \nabla_{\btheta} f = \bbv^\top G \hspace{0.2em} (\textrm{grad} f)$.
%
% \begin{align*}
%      df(\bV) &= \langle \bV, \textrm{grad} f\rangle  \quad \textrm{ by definition } df(\bV) = \bV(f) \\
%      % \bV(f) &= \langle \bMd \bbv, \bMd (\textrm{grad} f) \rangle \\
%      % \sum_i \bbv^i  \tfrac{\partial}{\partial \hspace{-0.04cm} \btheta_i} f
%      % &= \boldsymbol{v}^\top \boldsymbol{M}_{\partial}^\top \boldsymbol{M}_{\partial} (\textrm{grad} f) \\
%        \bbv^\top \nabla_{\btheta} f
%      &= \bbv^\top G \hspace{0.2em} (\textrm{grad} f).
% \end{align*}
%
Then we see $\nabla_{\btheta} f = G  \hspace{0.2em}(\textrm{grad} f)$ and $(\textrm{grad} f) = G^{-1} \hspace{0.2em} \nabla_{\btheta} f$. Note that $f(\bx) = x_{D+1}$ so that $f(\xi) = \ell $ which yields $\textrm{grad} \hspace{0.1em} f(\bx) = \bMd G^{-1} \nabla \ell$.
% %
% \begin{align}
%  \textrm{grad} \hspace{0.1em} f(\bx) =& \bMd G^{-1} \nabla \ell.
% \end{align}
% %
From where we identify the expression $G^{-1} \nabla \ell$ as the Natural gradient, which are the components of the gradient vector of $f$ at $\bx = \xi$. We also note that \citet{boumal:2023} provides an easier way to express the Riemmanian gradient on embedded manifolds.

% by defining it in the following way. Denote the extension of $f$ on the Euclidean space $\mathbb{R}^{D + 1}$ as $\tilde{f}$ such that $f = \tilde{f}|_{\Gamma_\ell}$. Apply the orthogonal projection of its classical gradient $\nabla \tilde{f}$ towards the tangent spaces of the embedded manifold $\Gamma_\ell \subset \nman$. The resulting operation gives $\textrm{grad}f(\bx)$. In our case, we would do it in two steps. First obtain the Riemannian gradient on the tangent space of $\nman$ and then project this gradient on the tangent space of $\Gamma_\ell$. The resulting operations give the same results. 
 
\section{Normal vector on $\Gamma_\ell$} \label{app:normalv}

This section computes the normal vector at a point $\bx \in \Gamma_\ell$ considering the warped metric $\langle \cdot, \cdot \rangle_{\psi}$. The notation $\bN_{\bx}$ or $\bN$ will be used interchangeably. Denote $\nman \ni \bZ = \bZ^T + \bN$ where $\bZ^T \in T_{\bx}\Gamma_\ell$ and $\bN$ is the normal vector to $T_{\bx}\Gamma_\ell$. Considering the canonical parametrisation $\xi$, the orthogonality under the metric $\langle \cdot, \cdot \rangle_{\psi}$ gives $\langle \partial_i \xi, \bN \rangle_{\psi} = 0$ for $i = 1, \ldots, D$. This implies the system of equations $\bN_i = - \psi^2 \bN_{D + 1} \partial_i \ell$. Assuming that the normal vector has unit length we have $\norm{\bN}_{\psi}^2 = 1$. Using the system of equations to solve for the coordinate $\bN_{D + 1}$ we get
$$
\sum_{i = 1}^D (\bN_i)^2 + \psi^2 (\bN_{D + 1})^2 = 1,
$$
and solving for the last coordinate we obtain $ \Big(\psi^4 \sum_{i = 1}^D (\partial_i \ell)^2 + \psi^2 \Big) (\bN_{D + 1})^2 = 1$. This leads to $ \bN_{D + 1} = \frac{1}{\psi \sqrt{\psi^2 \|\nabla \ell\|^2 + 1}}$.
Therefore the normal vector at $\bx \in \Gamma_\ell$ is
\begin{align}
\bN_{\bx} 
% & = \left( - \frac{\psi \partial_1 \ell}{\sqrt{\psi^2 \|\nabla \ell\|^2 + 1}}, \ldots, - \frac{\psi \partial_D \ell}{\sqrt{\psi^2 \|\nabla \ell\|^2 + 1}}, \frac{1}{\psi \sqrt{\psi^2 \|\nabla \ell\|^2 + 1}} \right) \nonumber \\
& = \left( -\frac{\psi\nabla \ell}{W}, \frac{1}{\psi W} \right)
\end{align}
where $W = \sqrt{\psi^2 \| \nabla \ell \|^2 + 1}$ and this vector has unit norm under the metric $\langle \cdot, \cdot \rangle_{\psi}$.

\section{Orthogonal projection on $T_{\bx}\Gamma_\ell$} \label{app:orthop}

Again denote $\nman \ni \bZ = \bZ^T + \bN$ where $\bZ^T \in T_{\bx}\Gamma_\ell$ and $\bN$ is the normal vector to $T_{\bx}\Gamma_\ell$. We known that $\langle \bZ^T, \bN \rangle_\psi = 0$ and we want to find the orthogonal projection of $\bZ$ onto $T_{\bx}\Gamma_\ell$. Since $\bZ^T \in T_{\bx}\Gamma_\ell$ we need to find the coordinate components $\bbv$ of $\bZ^T = \boldsymbol{M}_\partial\bbv$. To do so we need to solve $\langle \boldsymbol{M}_\partial \bbv, \bZ - \boldsymbol{M}_\partial \bbv \rangle_\psi = 0$ for $\bbv$. This is given by the weighted least-square solution. That is, 
$$
\bbv = \left( \boldsymbol{M}_\partial^\top
G_\psi \boldsymbol{M}_\partial \right)^{-1} \boldsymbol{M}_\partial^\top 
G_\psi 
\bZ
$$
where $G_\psi = \diag(I, \psi^2)$ is the metric-tensor of the ambient space $\nman$. Therefore the orthogonal projection of a vector $\bZ$ onto $T_{\bx}\Gamma_\ell$, denoted as $\textsl{Proj}_{T_{\bx}\Gamma_\ell} \bZ$ has the form,
$$
\textsl{Proj}_{T_{\bx}\Gamma_\ell} \bZ = \boldsymbol{M}_\partial \left( \boldsymbol{M}_\partial^\top 
\begin{bmatrix}
I & \0 \\ 
\0^\top & \psi^2
\end{bmatrix} 
\boldsymbol{M}_\partial \right)^{-1} \boldsymbol{M}_\partial^\top 
\begin{bmatrix}
I & \0 \\ 
\0^\top & \psi^2
\end{bmatrix} 
\bZ.
$$
Observe that $\boldsymbol{M}_\partial^\top \diag(I, \psi^2) \boldsymbol{M}_\partial = I_D + \psi^2 \nabla \ell \nabla \ell^\top$ is the metric-tensor induced on $\Gamma_\ell$. Expanding the terms above we get, 
\begin{align*}
    \textsl{Proj}_{T_{\bx}\Gamma_\ell} \bZ &= \boldsymbol{M}_\partial \big( I_D - \tfrac{\psi^2}{W^2} \nabla \ell \nabla \ell^\top \big ) [I_D \ \psi^2 \nabla \ell] \bZ \\
    % &= \boldsymbol{M}_\partial \big[I_D - \tfrac{\psi^2}{W^2} \nabla \ell \nabla \ell^\top \ \psi^2 \nabla \ell - \tfrac{\psi^4}{W^2} \| \nabla \ell \|^2 \nabla \ell \big] \bZ \\
    % &= \boldsymbol{M}_\partial \big[I_D - \tfrac{\psi^2}{W^2} \nabla \ell \nabla \ell^\top \ \psi^2 \nabla \ell - \tfrac{\psi^2(W^2 - 1)}{W^2} \nabla \ell \big] \bZ \\
    % &= \boldsymbol{M}_\partial \big[I_D - \tfrac{\psi^2}{W^2} \nabla \ell \nabla \ell^\top \ \tfrac{\psi^2}{W^2} \nabla \ell \big] \bZ \\
    % &= \boldsymbol{M}_\partial \Big( 
    %     \bZ_{1:D} - \tfrac{\psi^2}{W^2} \inp{\nabla \ell}{\bZ_{1:D}} \nabla \ell + \bZ_{D + 1} \tfrac{\psi^2}{W^2} \nabla \ell \Big) \\
    &= \boldsymbol{M}_\partial \Big[
        \bZ_{1:D} + \tfrac{\psi^2}{W^2} \Big(\bZ_{D + 1} - \inp{\nabla \ell}{\bZ_{1:D}} \Big) \nabla \ell \Big] \\
\end{align*}

\section{Christoffel symbols and geodesic equations on $\Gamma_\ell$} \label{app:chrisgeod}

Recall that $\psi$ is a function of $\bx$, the metric-tensor induced on $\Gamma_\ell$ is $G = I_D + \psi^2 \nabla \ell  \nabla \ell^\top$ whose inverse is $ G^{-1} = I_D - \tfrac{\psi^2}{W^2}\nabla \ell \nabla \ell^\top $ where $W = \sqrt{\psi^2 \|\nabla \ell\|^2 + 1}$. The Christoffel symbols $\Gamma^m_{i, j}$ for $i, j, m = 1, \ldots, D$ are computed using its general formulation \citep[see][for example]{docarmo:1992}. After some algebraic manipulation we can organize the Christofell symbols in matrices. The development leads to,
\begin{align*}
    \Gamma^m_{i, j} = 
    % & \tfrac{1}{2} \sum_{k = 1}^D G^{-1}_{k, m} \big(\partial_i G_{j, k} + \partial_j G_{k, i} - \partial_k G_{i, j} \big) \\
        % = & \tfrac{1}{2} \sum_{k = 1}^D \left( \delta_{k, m} - \tfrac{\psi^2}{W^2} \partial_k \ell \partial_m \ell \right)
        % \Big(\partial_i \psi^2  \partial_j \ell \partial_k \ell 
        % + 
        % {\color{darkgray} \psi^2 \partial^2_{i, j} \ell \partial_k \ell} 
        % + 
        % {\color{gray} \psi^2 \partial_j \ell \partial^2_{i, k} \ell} + \partial_j \psi^2 \partial_k \ell \partial_i \ell \\ 
        % {\color{white} =} & + 
        % {\color{gray} \psi^2 \partial^2_{j, k} \ell \partial_i \ell} 
        % + 
        % {\color{darkgray} \psi^2 \partial_k \ell  \partial^2_{j, i} \ell} 
        % - \partial_k \psi^2\partial_i \ell \partial_j \ell
        % - {\color{gray} \psi^2 \partial^2_{k, i} \ell \partial_j \ell} - {\color{gray}\psi^2 \partial_i \ell \partial^2_{k, j} \ell} \Big) \\[0.19cm]
        & \tfrac{1}{2} \sum_{k = 1}^D \left( \delta_{k, m} - \tfrac{\psi^2}{W^2} \partial_k \ell \partial_m \ell \right)
        \Big(\partial_i \psi^2  \partial_j \ell \partial_k \ell + 
        {\color{darkgray} 2 \psi^2 \partial^2_{i, j} \ell \partial_k \ell} + \partial_j \psi^2 \partial_k \ell \partial_i \ell - \partial_k \psi^2\partial_i \ell \partial_j \ell \Big).
\end{align*}
 Because the Hessian $\nabla^2 \ell$ is symmetric, some terms will cancel out and others combine. Expanding the summation above, we get 
\begin{align*}
    \Gamma^m_{i, j} 
    % & = \tfrac{1}{2} \bigg(\partial_i \psi^2 \partial_j \ell \partial_m \ell - \tfrac{\psi^2}{W^2} \partial_i \psi^2 \partial_j \ell \partial_m \ell \sum_{k = 1}^D \partial_k \ell^2 + \ldots \bigg) \\ 
    % & = \tfrac{1}{2} \bigg[ \bigg(1 - \frac{\psi^2}{W^2} \sum_{k = 1}^D \partial_k \ell^2 \bigg) \bigg(\partial_i \psi^2 \partial_j \ell \partial_m \ell + 2\psi^2 \partial^2_{i, j} \ell \partial_m \ell + \partial_j \psi^2 \partial_i \ell \partial_m \ell \bigg) + \ldots \bigg] \\
    & = \tfrac{1}{2} \bigg( \frac{1}{W^2} \partial_i \psi^2 \partial_j \ell \partial_m \ell + 
    \frac{2 \psi^2}{W^2} \partial_m \ell \partial^2_{i, j} \ell + 
    \frac{1}{W^2} \partial_j \psi^2 \partial_i \ell \partial_m \ell -
     \partial_m \psi^2 \partial_i \ell \partial_j \ell \\
    & \phantom{=} + 
    \frac{\psi^2}{W^2} \sum_{k = 1}^D \partial_k \psi^2 \partial_k \ell
    \partial_m \ell \partial_i \ell \partial_j \ell
    \bigg).
\end{align*}
Note that, except for the last term, all the terms in the first passage are computed similarly. That is why the equation is shortened. In the last passage we explicitly show the complete form of all the terms composing $\Gamma^m_{i, j}$. The Christoffel symbols when arranged in full matrices are denoted as $\Gamma^m$, $m = 1, \ldots, D$ and are generally written as,

\begin{align*}
    \Gamma^m = & \tfrac{1}{2} \bigg( \frac{1}{W^2} \nabla \psi^2  \nabla \ell^\top \partial_m \ell +
    \frac{1}{W^2} \nabla \ell (\nabla \psi^2)^\top \partial_m \ell + 
    \frac{2 \psi^2}{W^2} \nabla^2 \ell \partial_m \ell \\[0.2cm] 
    \phantom{=} & + 
    \frac{\psi^2 \langle \nabla \psi^2, \nabla \ell \rangle}{W^2} \nabla \ell \nabla \ell^\top \partial_m \ell - \nabla \ell \nabla \ell^\top \partial_m \psi^2 \bigg).
\end{align*}
To further simplify the notation, let
$$
\Lambda = \nabla \psi^2 \nabla \ell^\top + \nabla \ell (\nabla \psi^2)^\top + 2 \psi^2 \nabla^2 \ell + \psi^2 \langle \nabla \psi^2, \nabla \ell \rangle \nabla \ell \nabla \ell^\top. 
$$
Thus,
\begin{align*}
    \Gamma^m = \tfrac{\Lambda}{2 W^2} \partial_m \ell - \tfrac{1}{2} \nabla \ell \nabla \ell^\top \partial_m \psi^2.
\end{align*}

The computation of the geodesic equations associated to $\Gamma_\ell$ will also follow the general formalism. We will use the results above to make the equations more compact aiming at computational purposes. Using the geodesic equations and 
% %
% \begin{align*}
%     \dot \bbv = - \begin{bmatrix}
%     \norm{\bbv}^2_{\Gamma^1(\gamma(t))} \\ 
%     \vdots \\ 
%     \norm{\bbv}^2_{\Gamma^D(\gamma(t))}
%     \end{bmatrix}.
% \end{align*}
% %
expanding one element in its right hand-side, Equation \eqref{eq:geodeq}, we get,
\begin{align*}    
    \norm{\bbv}^2_{\Gamma^m(\gamma(t))} & = 
    \boldsymbol{v}^\top \big( \tfrac{\Lambda}{2 W^2} \partial_m \ell - \tfrac{1}{2} \nabla \ell \nabla \ell^\top \partial_m \psi^2 \big) \bbv.
\end{align*}
Observe that the quadratic form $\|\boldsymbol{v}\|^2_{\Lambda}$ can be expanded to have a computational-friendly expression, so that we do not need to form squared matrices. It follows, 
\begin{align*}
    \mho_1 = \tfrac{1}{2 W^2} \boldsymbol{v}^\top \Lambda \bbv =& \tfrac{1}{2 W^2} \Big( 2 \langle \bbv, \nabla \psi^2 \rangle \langle \bbv, \nabla \ell \rangle + 2 \psi^2 \| \boldsymbol{v} \|^2_{\nabla^2 \ell} + \psi^2 \langle \nabla \psi^2, \nabla \ell \rangle \langle \bbv, \nabla \ell \rangle^2 \Big) \\
    =& \tfrac{1}{W^2} \langle \bbv, \nabla \psi^2 \rangle \langle \bbv, \nabla \ell \rangle + \tfrac{\psi^2}{W^2} \| \boldsymbol{v} \|^2_{\nabla^2 \ell} + \tfrac{\psi^2}{2 W^2} \langle \nabla \psi^2, \nabla \ell \rangle \langle \bbv, \nabla \ell \rangle^2
\end{align*}
and
\begin{align*}
    \mho_2 = \tfrac{1}{2} \boldsymbol{v}^\top \nabla \ell \nabla \ell^\top \bbv = \tfrac{1}{2} \langle \bbv , \nabla \ell \rangle^2
\end{align*}
We now clearly see the need of only two gradients, each of which are  multiplied respectively by $\mho_1$ and $\mho_2$ which are scalar numbers. Therefore the geodesic equations above become, 
\begin{align} \label{eq:geodequations}
    \dot \bbv = - \mho_1(\bx, \bbv) \nabla \ell(\bx) + \mho_2(\bx, \bbv) \nabla \psi^2(\bx)
\end{align}
indeed all elements of the above equation are dependent on $t$.

\section{Second-fundamental form on $T_{\bx}\Gamma_\ell$} \label{app:seconf}

The second-fundamental form acting on the tangent space of $\Gamma_\ell$ in computed as follows. In short notation the normal vector on the ambient space will be denoted $\bar{\bN}  = \bar{\bMd} \bN$, where $\bN = \Big(- \tfrac{\psi\nabla \ell}{W}, \tfrac{1}{\psi W} \Big)$. Its Euclidean partial derivative in the $i^{th}$ direction $\partial_i \bar{\bN}$ is given by
$$
\partial_i \bar{\bN} = \Big[- \big(\partial_i \psi \tfrac{1}{W} + \psi \partial_i\tfrac{1}{W} \big) \nabla \ell - \tfrac{\psi}{W} \nabla^2 \ell_i,  \partial_i \tfrac{1}{\psi} \tfrac{1}{W} + \tfrac{1}{\psi} \partial_i \tfrac{1}{W}\Big].
$$
for $i = 1, \ldots, D$ and $\partial_{D + 1} \bar{\bN} = \0$. The matrix $\bar{\bMd} = I_{D+1}$, it is composed by the canonical basis of $\mathbb{R}^{D+1}$. The second-fundamental form $\mathbb{II}$ in the direction of $\bV = (\bbv, \inp{\bbv}{\nabla \ell})$ is defined as,
\begin{align*}
    \mathbb{II}_{\bx}(\bV) = - \inp{\bar{\nabla}_{\bar{\bV}} \bar{\bN}}{\bar{\bV}}_{\psi} 
    % = - \inp{\bar{\nabla}_{\bV} \bN}{\bV}_{\psi}
\end{align*}
where $\bar{\bV} = \bar{\bMd} \bV$,
% \footnote{Observe that the extension $\bar{\bV}$ on $\nman$ is the same as $\bV$}
$\bx = (x_1, \ldots, x_{D + 1}) \in \nman$ and  $\bar{\nabla}$ is the connection associated with the warped metric in the ambient space $\nman = \mathbb{R}^{D + 1}$. In the ambient space, the Christoffel symbols  $\bar{\Gamma}^m$ (in matrices forms) associated with the connection $\bar{\nabla}$ are given by
$$
\bar{\Gamma}^m = \tfrac{1}{2} \diag(\0, -\partial_m \psi^2)
$$
for $m = 1, \ldots, D$ and 
$$
\bar{\Gamma}^{D+1} = \tfrac{1}{2\psi^2} 
\begin{bmatrix}
 \0 & \nabla \psi^2 \\ 
 \nabla \psi^{2\top} & \0 
 \end{bmatrix}
$$
for $m = D + 1$ since $\psi$ does not depend on $x_{D + 1}$. The covariant derivative $\bar{\nabla}_{\bar{\bV}} \bar{\bN}$ can be computed using the general definition in \cite{docarmo:1992}. It follows,
\begin{align*}
\bar{\nabla}_{\bar{\bV}} \bar{\bN} = &
\bar{\bMd} \begin{bmatrix}
 \bar{\bV}(\bN_1) + \inp{\bV}{\bN}_{\bar{\Gamma}_1}\\ 
 \vdots\\ 
 \bar{\bV}(\bN_{D + 1}) + \inp{\bV}{\bN}_{\bar{\Gamma}_{D + 1}}
 \end{bmatrix} \\ 
 =& \bar{\bMd} \left( \sum_{i = 1}^{D + 1} \bar{\boldsymbol{v}}_i \partial_i \bN + \left(- \tfrac{1}{2}\tfrac{1}{\psi W} \inp{\bbv}{\nabla \ell} \nabla \psi^2, \tfrac{1}{2\psi^2} \big( - \tfrac{\psi}{W} \inp{\bbv}{\nabla \ell} \inp{\nabla \psi^2}{\nabla \ell} + \tfrac{1}{\psi W} \inp{\bbv}{\nabla \psi^2} \big) \right) \right).
\end{align*}
where $\bar{\boldsymbol{v}}_i$ is the coordinate of $\bar{\bV}$ and $\bar{\boldsymbol{v}}_i = \boldsymbol{v}_i$ for $i = 1, \ldots, D$. Observe that $\bar{\boldsymbol{v}}_{D+1} = \inp{\nabla \ell}{\boldsymbol{v}}$, however $\partial_{D + 1} \bar{\bN} = \0$ which makes the last term in the sum above disappear. Plugging the covariant derivative $\bar{\nabla}_{\bV} \bar{\bN}$ and the tangent vector $\bar{\bV}$ into the definition of the second-fundamental form yields 
{\small
\begin{align*}
\inp{\bar{\nabla}_{\bar{\bV}} \bar{\bN}}{\bar{\bV}}_{\psi} &= \left\langle \bar{\bMd} \sum_{i = 1}^D \boldsymbol{v}_i \partial_i \bN, \ \bar{\bMd}(\bbv, \inp{\bbv}{\nabla \ell})
\right\rangle_\psi \\
&+ \left\langle \bar{\bMd} \big( - \tfrac{1}{2\psi W} \inp{\bbv}{\nabla \ell} \nabla \psi^2, - \tfrac{1}{2\psi W} \inp{\bbv}{\nabla \ell} \inp{\nabla \psi^2}{\nabla \ell} + \tfrac{1}{2\psi^3 W} \inp{\bbv}{\nabla \psi^2} \big), \ \bar{\bMd} \big(\bbv, \inp{\bbv}{\nabla \ell} \big) \right\rangle_{\psi}.
\end{align*}
}
After some algebraic manipulation the first term of the above sum becomes
{\small
\begin{align*}
    \left\langle \bar{\bMd} \sum_{i = 1}^D \boldsymbol{v}_i \partial_i \bN, \bar{\bMd} (\bbv, \inp{\bbv}{\nabla \ell}) \right\rangle_\psi =& 
    \bbv^\top \Big[ - \Big( \nabla \psi \tfrac{1}{W} + \psi \nabla \tfrac{1}{W} \Big) \nabla \ell^\top - \tfrac{\psi}{W} \nabla^2 \ell + \psi^2 \Big( \nabla \tfrac{1}{\psi} \tfrac{1}{W} + \tfrac{1}{\psi} \nabla \tfrac{1}{W} \Big) \nabla \ell^\top \Big] \bbv \\
    % =& \bbv^\top \Big( - \tfrac{2}{W} \nabla \psi \nabla \ell^\top + \psi \nabla \tfrac{1}{W} \nabla \ell^\top - \psi \nabla \tfrac{1}{W} \nabla \ell^\top - \tfrac{\psi}{W} \nabla^2 \ell  \Big)  \bbv \\
    =& \bbv^\top \Big( - \tfrac{2}{W} \nabla \psi \nabla \ell - \tfrac{\psi}{W} \nabla^2 \ell  \Big) \bbv.
\end{align*}
}
Therefore, considering the negative sign, the second-fundamental acting on the tangent space of $\Gamma_\ell$ is given by
\begin{align*}
\mathds{II}_{\bx}(\bV) &= - \inp{\bar{\nabla}_{\bar{\bV}} \bar{\bN}}{\bar{\bV}}_{\psi} \\
% &= - \bbv^\top \Big( - \tfrac{2}{W} \nabla \psi \nabla \ell^\top - \tfrac{\psi}{W} \nabla^2 \ell - \tfrac{\psi}{2 W} \inp{\nabla \psi^2}{\nabla \ell} \nabla \ell \nabla \ell^\top \big) \bbv \\
% &= \bbv^\top \Big( \tfrac{2}{W} \nabla \psi \nabla \ell^\top + \tfrac{\psi}{W} \nabla^2 \ell + \tfrac{\psi}{2 W} \inp{\nabla \psi^2}{\nabla \ell} \nabla \ell \nabla \ell^\top \Big) \bbv \\
&= \tfrac{2}{W} \langle \bbv, \nabla \psi \rangle \langle \bbv, \nabla \ell \rangle + \tfrac{\psi}{W} \| \boldsymbol{v} \|^2_{\nabla^2 \ell} + \tfrac{\psi}{2 W} \langle \nabla \psi^2, \nabla \ell \rangle \langle \bbv, \nabla \ell \rangle^2
\end{align*}

\section{Third-order Taylor expansion of the geodesic curve} \label{app:taylor3}

The third-order order degree Taylor approximation of the geodesic curve on a point $\bx \in \Gamma_\ell$ in the direction of $\bV \in T_{\bx}\Gamma_\ell$ is given by 
\begin{align}
	\tilde{\gamma}_{\bx, \bV}(t_*) = \bx + t_* \boldsymbol{V} + \frac{t_*^2}{2} \ddot \gamma(0) + \frac{t_*^3}{6} \dddot \gamma(0).
\end{align}
where $\ddot \gamma(0) = Q_{\bx}(\bV)$ is the normal component of the geodesic curve on $\nan \times \man_{\psi}$. This normal component is given by the second-fundamental form multiplied by the normal vector since geodesics have null tangential component. That is, for a geodesic $\gamma(t)$, $\ddot \gamma(t) = Q_{\gamma(t)}(\dot \gamma(t)) = \mathbb{II}_{\gamma(t)}(\dot \gamma(t)) \bar{\bN}_{\gamma(t)}$. Expanding this expression at $t = 0$ we get
\begin{align}
Q_{\bx}(\bV) = \big( \tfrac{2}{W} \inp{\bbv}{\nabla \psi} \inp{\bbv}{\nabla \ell} + \tfrac{\psi}{W} \|\boldsymbol{v}\|^2_{\nabla^2 \ell} + \tfrac{\psi}{2 W} \inp{\nabla \psi^2}{\nabla \ell} \inp{\nabla \ell}{\bbv}^2  \big) 
\begin{bmatrix}
- \tfrac{\psi\nabla \ell}{W} \\
\tfrac{1}{\psi W}
\end{bmatrix}
\end{align}
where for a given $\bV$ the coordinates components $\bbv$ can be recovered using the orthogonal projection above, that is, $\bbv = \big( \boldsymbol{M}_\partial^\top G_\psi \boldsymbol{M}_\partial \big)^{-1} \boldsymbol{M}_\partial^\top G_\psi \bV$. The third-order component of the approximate geodesic is $\dddot \gamma(0) = K_{\bx}(\bV)$ and obtained by taking the time derivative of $\ddot \gamma(t)$ at $t = 0$ in the direction of $\bbv$.
\begin{align}
\dddot \gamma(t) &= \frac{\mathrm{d}}{\mathrm{d}t} Q_{\gamma(t)}(\dot \gamma(t)) \\ \nonumber
 &= \frac{\mathrm{d}}{\mathrm{d}t} \bigg( \tfrac{2}{W} \inp{\bbv}{\nabla \psi} \inp{\bbv}{\nabla \ell} + \tfrac{\psi}{W} \|\boldsymbol{v}\|^2_{\nabla^2 \ell} + \tfrac{\psi}{2 W} \inp{\nabla \psi^2}{\nabla \ell} \inp{\nabla \ell}{\bbv}^2 \bigg) \begin{bmatrix}
- \tfrac{\psi\nabla \ell}{W} \\
\tfrac{1}{\psi W} 
\end{bmatrix}.
 \end{align}
Here we are interested in the first $D$ components of this approximation since $\tilde{\gamma}_{\bx, \bV}(t_*) \notin \Gamma_\ell$ usually. Following the retraction map choice, we apply the orthogonal projection of $\dddot \gamma$ towards $\mathcal{N} = \Theta$ to obtain its first $D$ component. Then we can write 
\begin{align}
\dddot \gamma(t)_{1:D} &= \frac{\mathrm{d}}{\mathrm{d}t} Q_{\gamma(t)}(\dot \gamma(t))_{1:D} \nonumber \\
% &= \frac{\mathrm{d}}{\mathrm{d}t} \big( \tfrac{2}{W} \inp{\bbv}{\nabla \psi} \inp{\bbv}{\nabla \ell} + \tfrac{\psi}{W} \|\boldsymbol{v}\|^2_{\nabla^2 \ell} + \tfrac{\psi}{2 W} \inp{\nabla \psi^2}{\nabla \ell} \inp{\nabla \ell}{\bbv}^2 \big) \big[ -\tfrac{\psi\nabla \ell}{W} \big] \nonumber \\
&= - \frac{\mathrm{d}}{\mathrm{d}t} \big( \tfrac{1}{W^2} \inp{\bbv}{\nabla \psi^2} \inp{\bbv}{\nabla \ell} + \tfrac{\psi^2}{W^2} \|\boldsymbol{v}\|^2_{\nabla^2 \ell} + \tfrac{\psi^2}{2 W^2} \inp{\nabla \psi^2}{\nabla \ell} \inp{\nabla \ell}{\bbv}^2 \big) \nabla \ell 
% &= - \frac{\mathrm{d}}{\mathrm{d}t} \big( \mho_1 \nabla \ell \big)
\end{align}
where we have used that $\nabla \psi^2 = 2 \psi \nabla \psi$. The particular derivatives which compose the complete time derivative above are straightforward to compute and accordingly to the choice of the warp function.

\section{The choice of warp function} \label{app:warpf}

\begin{figure}[!t]
    \centering
    \includegraphics[width = 12.5cm, height = 4.85cm]{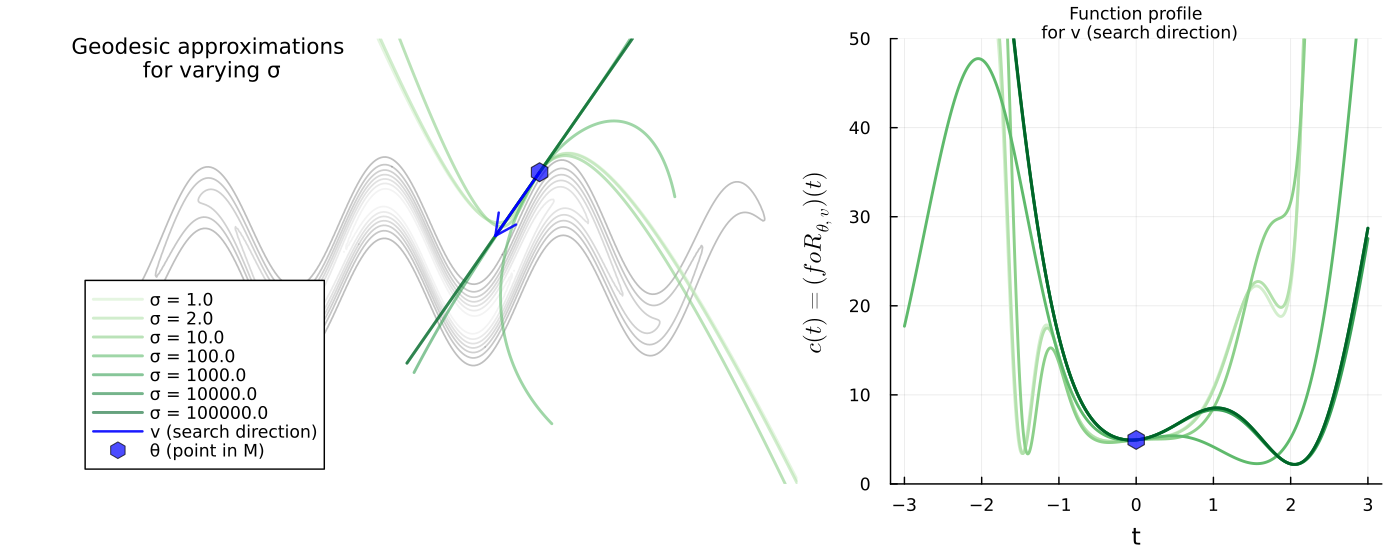}
    \caption{Geodesic approximations based on $3^{\textrm{rd}}$-order Taylor series. The level set in gray represents the function $\ell(\btheta) = \log \mathcal{N}\big([\theta_1, \theta_2 + \sin(1.3 \theta_1)]|\bmu, \Sigma \big)$ where $\mathcal{N}$ denotes the Gaussian density  $\bmu = \0$ and $\Sigma = \diag(20, 0.1)$. The blue point $\xi^{-1}(\bx) = [3.0 \ 1.4]^\top$ and blue vector $\bbv = [-1.2 \ -1.0]^\top$ display the point and direction where the approximation of the geodesic curve on is made for a series of increasing $\sigma^2$ values. As $\sigma^2$ values increase the approximations tend to be closer to a straight line and in the limit of $\sigma^2 \rightarrow \infty$ the geodesic approximation becomes aligned with the search direction $\bbv$.}
    \label{fig:geodcurv}
\end{figure}

Suppose that we have a natural embedding of $\Gamma_\ell$ on $\mathbb{R}^{D + 1}$ equipped with the Euclidean metric and the canonical parametrisation $\xi$ aforementioned. Then the normal vector at $\bx \in \Gamma_\ell \subset \mathbb{R}^{D + 1}$ is given by $
\bN_* = \big(- \nabla \ell / \sqrt{\| \nabla \ell \|^2 + 1}, 1 / \sqrt{\| \nabla \ell \|^2 + 1} \big).
$
% $
% \bN_* = \left(- \tfrac{\nabla \ell}{\sqrt{\| \nabla \ell \|^2 + 1}}, \tfrac{1}{\sqrt{\| \nabla \ell \|^2 + 1}} \right).
% $
%
We first define the warp function to be the norm of the orthogonal projection of $\bN_*$ over $T_{\xi^{-1}(\bx)}\Theta$. This implies that $\psi_* = \|\textsl{Proj}_{T_{\xi^{-1}(\bx)}\Theta}\bN_* \| = \| \nabla \ell \|/ (\sqrt{1 + \| \nabla \ell \|^2})$.
% %
% \begin{align*}
%    \psi_* & = \|\textsl{Proj}_{T_{\xi^{-1}(\bx)}\Theta}\bN_* \| = \frac{\| \nabla \ell \|}{\sqrt{1 + \| \nabla \ell \|^2}}.
% \end{align*}
% %
From here we can see that $\psi_* \in (0, 1)$. We can see that in regions far away from the optima of $\ell$, the function $\psi_* \rightarrow 1^{-}$ as we will expect the components of the gradient $\nabla \ell$ to have high magnitude. Note that close to the optima $\psi_* \rightarrow 0^+$ as the gradient components tend to zero and the metric $G = I$ (identity), so that at the optima we recover the Euclidean metric. However, we believe that this function may be too restrictive for functions $\ell$ that can induce strong "bending" on the approximate geodesic path which may undesirable for practical purposes. For this reason we propose a more flexible warp function defining
\begin{align}
   \psi = \frac{ \alpha \| \nabla    \ell \|}{\sqrt{ \sigma^2 + \| \nabla \ell \|^2}},
\end{align}
where $\alpha, \sigma^2 > 0$ are scalars controlling respectively the upper bound and the flattening of the function $\psi \in (0, \alpha)$. The larger the values of $\sigma^2$ the smaller the function $\psi$ will be. If $\sigma^2 \rightarrow 0^+$ the function $\psi = \alpha$. If $\alpha = 1$ and $\sigma^2 = 0$, $\nan \times \man_\psi = \mathbb{R}^{D + 1}$. In order to visualize the behaviour of the $3^{\textrm{rd}}$ Taylor series in the approximation of geodesics with varying $\sigma$ and $\alpha = 1$ see Figure \ref{fig:geodcurv}. In this figure the example constructed considers the function $\ell(\btheta) = \log \mathcal{G}\big([\theta_1, \theta_2 + \sin(1.3 \theta_1)]|\0, \Sigma \big)$ where $\mathcal{G}$ denotes the Gaussian density function in two dimensions with $\bmu = \0$ and $\Sigma = \diag(20, 0.1)$. The approximations are made at $\xi^{-1}(\bx) = [3.0 \ 1.4]^\top$ and $\bbv = -[1.2 \ 1.0]^\top$ and both are depicted in blue colour. The panel displays the $3^{\textrm{rd}}$-order Taylor series approximation with a series of increasing $\sigma$ values. See also Equation \eqref{eq:geodret} for the general approximate geodesic expression. On the right side of the picture we plot the profile of the composite function $g(t)$ as a function of a scalar $t$. As $t$ varies, the plot on the right corresponds a walk-through along the approximate geodesics path on the left plot for different $\sigma^2$ values.    

In the previous calculations across the paper the gradient of the warp function $\psi^2$ was required. For this particular choice of warp function we obtain its gradient and time derivative as follows. Denote $W_{\sigma}$ $=$ $\sqrt{\sigma^2 + \| \nabla \ell \|^2}$, we have
\begin{align}
    \nabla \psi^2 & 
    % \alpha^2 \left( \nabla \| \nabla \ell \|^2 \frac{1}{W_{\sigma}^2} + \| \nabla \ell \|^2 \nabla \frac{1}{W_{\sigma}^2} \right) \nonumber 
    % & = \alpha^2 \left( 2 \nabla^2 \ell \nabla \ell \frac{1}{W_{\sigma}^2} - 2 \| \nabla \ell \|^2 \frac{1}{W_\sigma^4} \nabla^2 \ell \nabla \ell \right) \nonumber \\
    % & = \alpha^2 \left(\frac{2}{W_\sigma^2} - \frac{2 \| \nabla \ell \|^2}{W_\sigma^4} \right) \nabla^2 \ell \nabla \ell \nonumber 
    % & = \alpha^2 \left(\frac{2}{W_\sigma^2} - \frac{2 (W^2_\sigma - \sigma^2)}{W_\sigma^4} \right) \nabla^2 \ell \nabla \ell \nonumber \\
    = \frac{2 \alpha^2 \sigma^2}{W_\sigma^4} \nabla^2 \ell \nabla \ell, \ \ \ \     \frac{\mathrm{d}}{\mathrm{d}t} \nabla \psi^2 = \dfrac{2 \alpha^2 \sigma^2 }{W_\sigma^4} \left(-\dfrac{4}{W_\sigma^2} \inp{\bbv}{\nabla^2\ell \nabla \ell} \nabla^2 \ell \nabla \ell + \frac{\mathrm{d}}{\mathrm{d}t} \nabla^2 \ell \nabla \ell \right).
\end{align}
%
% and the time derivative of $\nabla \psi^2$ is given by
% %
% \begin{align} \label{eq:dtH}
%     \frac{\mathrm{d}}{\mathrm{d}t} \nabla \psi^2 = \dfrac{2 \alpha^2 \sigma^2 }{W_\sigma^4} \left(-\dfrac{4}{W_\sigma^2} \inp{\bbv}{\nabla^2\ell \nabla \ell} \nabla^2 \ell \nabla \ell + \frac{\mathrm{d}}{\mathrm{d}t} \nabla^2 \ell \nabla \ell \right).
% \end{align}

\section{Computational costs in the experiments} \label{app:comp_cost}

\begin{figure}[!t]
    \centering
    \subfloat[]{\includegraphics[scale = 0.18]{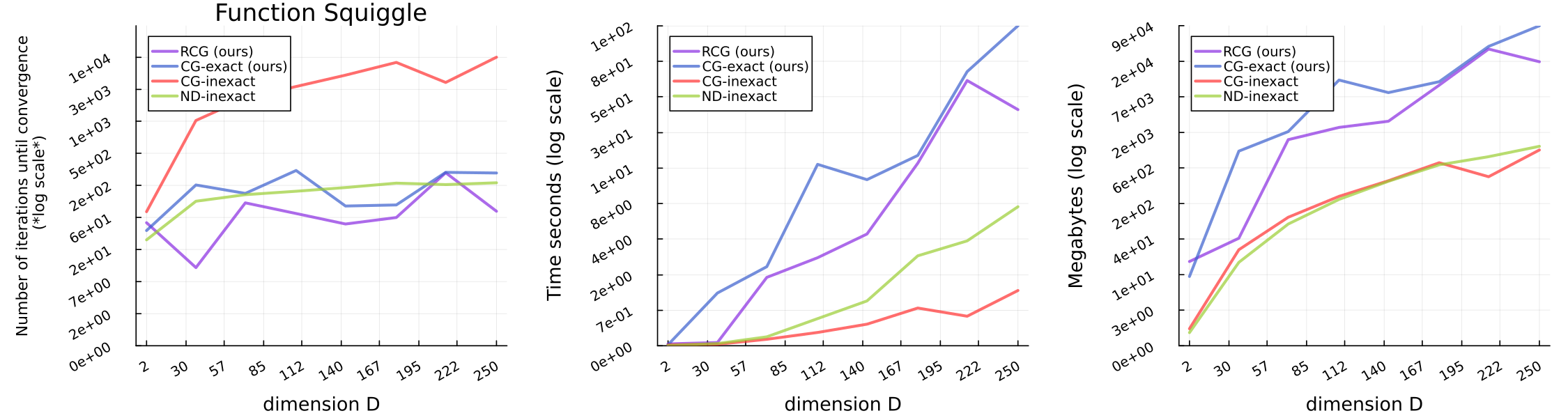}} \\
    \subfloat[]{\includegraphics[scale = 0.18]{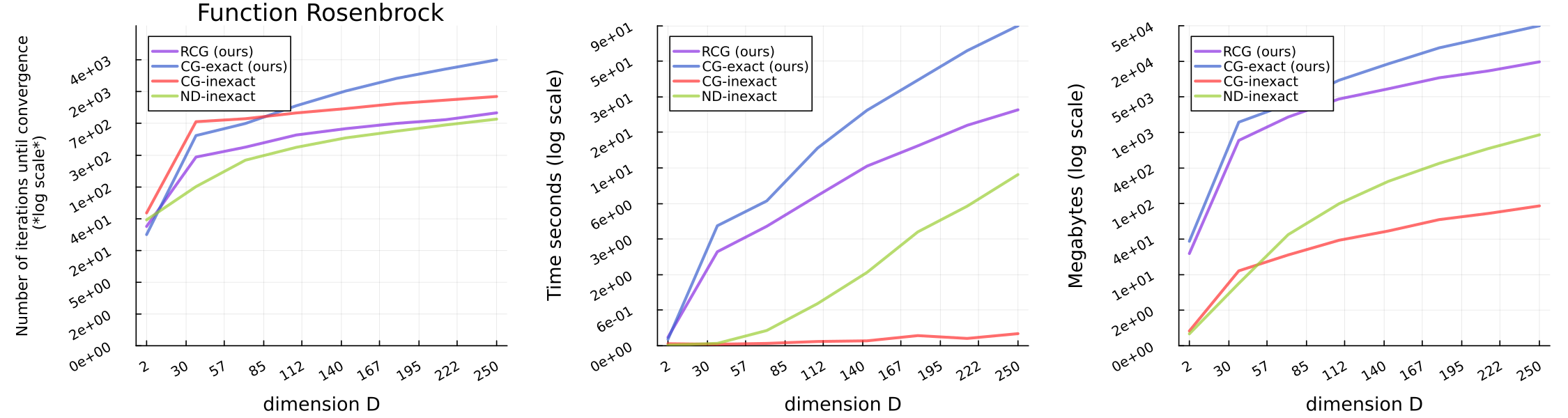}} \\
    \caption{In the panels (a) and (b) display the computation performance for the first two sets of models. The squiggle model and Rosenbrock model. The experiments are in terms of number of iterations, wall-clock time and memory consumption (from left to right). All measured until the stopping criteria of the the algorithms. Panel (a) shows the performance of all algorithms for the squiggle model. Panel (b) shows the same experiment but for the rosenbrock model. The performance of the RCG (ours) clearly improves the number of iterations until convergence, however the time and memory consumption until convergence has been higher than the state-of-art implementation of classical algorithms such as CG-inexact and ND-inexact in Julia language.}
    \label{fig:compcost}
\end{figure}

Here we show extra experiments showing the performance of RCG (ours) method proposed for the first two sets of models. In Figure \ref{fig:compcost}, we added the wall-clock time and memory consumption. Panel (a) shows the squiggle model. In this case ND-inexact, RCG (ours) and CG-exact (ours) dominate the speed of convergence. Time and memory comsumptions are greater for CG-exact (ours) when compared to other optimisers. For the Rosenbrock model, in panel (b), we see that RCG (ours) is faster in terms of number of iterations when compared to the to CG counterparts and closer to the ND-inexact. The time and memory consumption is dominated by the CG-exact (ours).

\bibliographystyle{rss}
\bibliography{refs}

\end{document}